\documentclass[10pt,twocolumn,letterpaper]{article}
\PassOptionsToPackage{table}{xcolor}
\usepackage[pagenumbers]{cvpr} 
\usepackage{graphicx}
\usepackage{booktabs}
\usepackage{amsmath}
\usepackage{cuted}
\usepackage{microtype}
\definecolor{cvprblue}{rgb}{0.21,0.49,0.74}
\usepackage[pagebackref,breaklinks,colorlinks,allcolors=cvprblue]{hyperref}

\def\paperID{*****} 
\def\confName{CVPR}
\def\confYear{2027}

\title{ESTHER: Egocentric Stereo Hand Estimation and Reconstruction in the Wild}
\author{Hongyu Ma, Hairong Qu, Shiqi Zhao, Yongsong Yang, Peng Yin\\
City University of Hong Kong}

\begin{document}

\maketitle

\begin{abstract}
Human dexterity is guided by two eyes watching two hands: binocular vision
supplies the metric 3D structure needed for precise manipulation. Egocentric stereo is therefore the natural perceptual
interface for robots, AR, and VR. Yet egocentric stereo hand reconstruction still lacks an end-to-end model
and an in-the-wild benchmark. We propose
ESTHER, a model whose
stereo geometry, temporal reasoning, and output representation are designed
for wearable egocentric stereo. It is trained on pseudo labels from a
calibrated labeling pipeline and in turn assembles our benchmark
ESTHER3D, an egocentric stereo hand dataset pairing a large in-the-wild
training set of
model labels with a motion capture test set of true metric
ground truth. Experiments show state-of-the-art accuracy, superior external
generalization across stereo rigs and modalities with minimal fine-tuning,
and robust metric reconstruction under missing views, monocular input,
dropped frames, and lighting and motion blur in real egocentric capture. \href{https://fruitrabbit.github.io/esther/}{Project page.}
\end{abstract}

\section{Introduction}

An embodied agent that sees the world through head-mounted stereo
perceives its own hands much as people do: two eyes watch two hands, and binocular disparity carries the metric 3D structure that precise manipulation depends on. This makes egocentric stereo the natural signal
for hand perception in robotics, AR, and VR: it provides the metric
constraints that monocular RGB lacks. We study this setting as
egocentric stereo 3D hand reconstruction: given short-baseline head-mounted
stereo videos, the system should recover MANO hand pose and shape,
and global wrist translation in metric 3D. Yet despite its practical
importance, egocentric stereo hand reconstruction still lacks an end-to-end
model and an in-the-wild benchmark. We identify two primary obstacles: the lack of
scalable training data with reliable metric hand labels, and the mismatch
between existing hand pose architectures and wearable stereo geometry.

\begin{figure}[t]
\centering
\includegraphics[width=0.82\columnwidth]{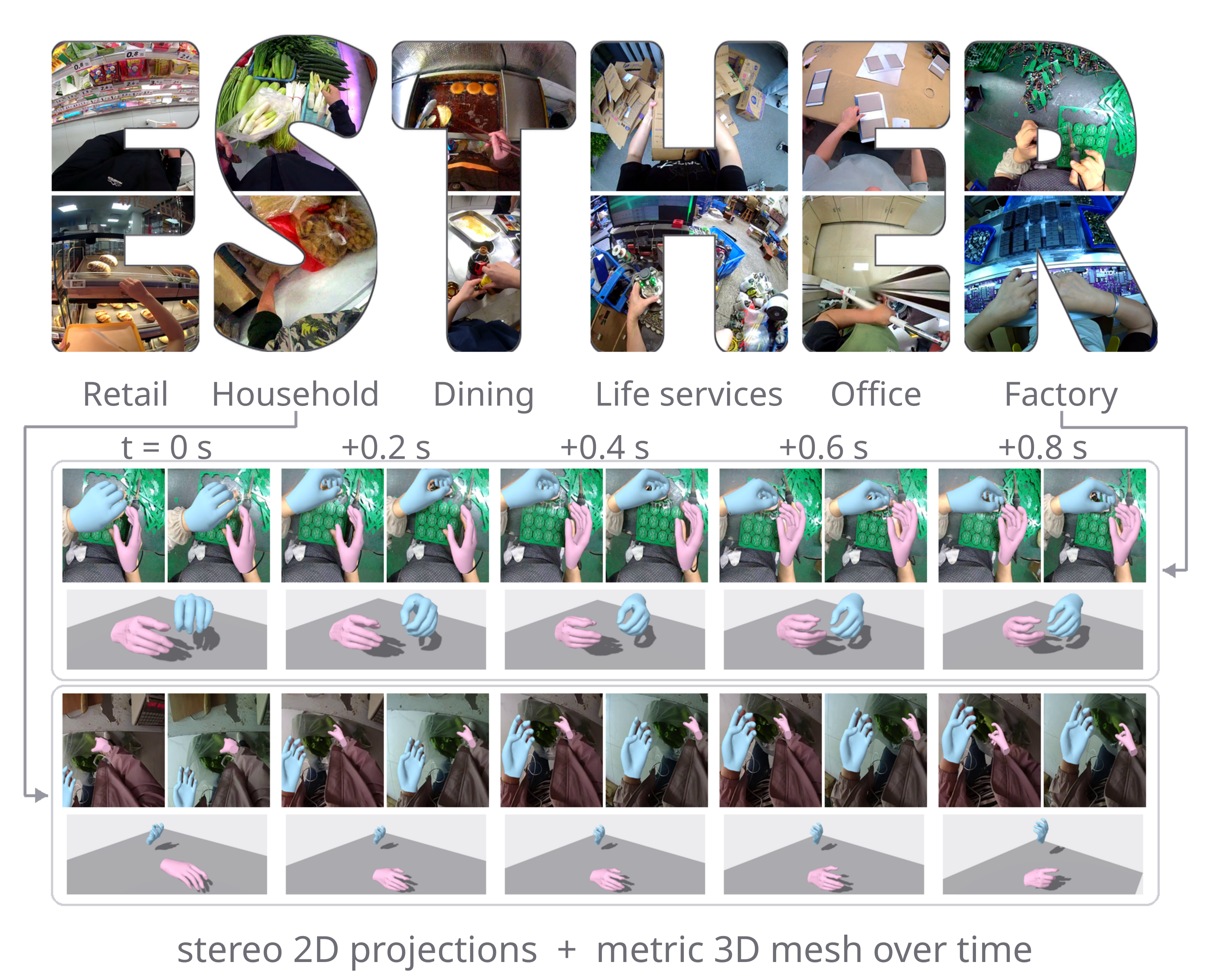}
\caption{\textbf{ESTHER at a glance.} Letters: egocentric stereo frames from
the six ESTHER3D industries. Bottom: ESTHER outputs stereo 2D projections and metric 3D meshes over time.}
\label{fig:teaser}
\end{figure}

Accurate 3D hand annotation for in-the-wild egocentric capture is hard: the
label itself must be recovered from partial, moving, and frequently occluded
observations. Existing egocentric hand datasets, including
H2O~\cite{h2o2021}, AssemblyHands~\cite{assemblyhands2023},
OakInk~\cite{oakink2022}, ARCTIC~\cite{arctic2023},
HOI4D~\cite{hoi4d2022}, HOT3D~\cite{hot3d2024}, obtain labels through
additional sensing or offline reconstruction; HOT3D is closest in
viewpoint, but its stereo is monochrome and its ground truth comes from
professional motion capture rather than scalable visual labeling of raw
RGB stereo. How to scale metric hand labels from raw in-the-wild
head-mounted stereo remains open. We therefore build a calibrated labeling
pipeline combining 2D evidence, stereo geometry, and hand priors, validate
it against motion capture ground truth, and train a scalable reconstruction
model on the pseudo labels the pipeline generates; the resulting benchmark,
ESTHER3D, pairs a mocap ground truth test set with these in-the-wild
training sequences.

The model bottleneck can be traced through the evolution of hand pose
architectures. Monocular mesh methods~\cite{hamer2024,wilor2025} regress
plausible hands, but from a single crop they can recover depth and scale only
through learned priors rather than geometric measurement, so metric accuracy
is not guaranteed.
Multiview
methods~\cite{learnabletriangulation2019,poemv2_2024,mvgformer2023}
restore scale through calibrated triangulation, but are single-frame and
assume at least two cameras see each point. Temporal models add video context in camera coordinates~\cite{htt2023,handformer2024} or lift monocular video into
world space with SLAM~\cite{hawor2025,dynhamr2025}, recovering the camera
only up to the scale ambiguity of monocular SLAM and offline optimization.
Egocentric stereo sharpens all three weaknesses at once.
Triangulation anchors are not guaranteed: blur and occlusion
routinely erase the hand from one of the two views, so a model dependent on triangulation loses its metric anchor exactly on the frames that
matter. Detection dropout is the norm: the interacting hand is
repeatedly missed for entire frames, so architectures that process each frame independently cannot run
stably without temporal inpainting. Hand and head motion are
coupled: observed image motion mixes articulation with head egomotion and
crop reparameterization, which a model without explicit disentanglement
absorbs into finger pose. The setting therefore requires a model designed
for these failure modes.

We address this setting with ESTHER, built from two encoders and a
decoder. A stereo image encoder fuses calibrated stereo evidence into
joint tokens; a masked temporal motion encoder reasons over these
tokens across time, with camera and crop motion supplied as conditioning
tokens; and a query-gated fusion decoder reads both encoders through
structured pose, shape, and wrist tokens and outputs MANO pose,
shape shared over time, and a direct metric wrist translation.

In summary, our contributions are as follows:
\begin{itemize}[leftmargin=*,topsep=2pt,itemsep=1pt]
    \renewcommand{\labelitemi}{$\bullet$}
    \item We propose ESTHER, an end-to-end egocentric stereo hand
    reconstruction model built from two encoders and a query-gated fusion
    decoder that fuses calibrated stereo geometry and temporal context to
    output metric MANO pose, shape, and an absolute wrist, with state-of-the-art accuracy and robustness.
    \item We introduce ESTHER3D, an in-the-wild egocentric stereo hand
    benchmark whose test set carries motion capture ground truth and whose
    training set consists of a large collection of stereo videos labeled by ESTHER,
    bootstrapped from a calibrated pipeline validated on that ground truth.
    \item We evaluate under a controlled protocol (shared backbone, data,
    and losses) on the ESTHER3D test set, on HOT3D, and with missing views and dropped frames, and uncover a binding effect between size and depth:
    short-baseline stereo guidance lets direct wrist regression internalize
    a prior tying apparent hand scale to metric depth, giving surprising
    accuracy when using the monocular fallback and cheap adaptation to different baselines,
    camera parameters, and even the grayscale modality.
\end{itemize}

\section{Related Works}

\begin{figure*}[t]
\centering
\includegraphics[width=0.76\textwidth]{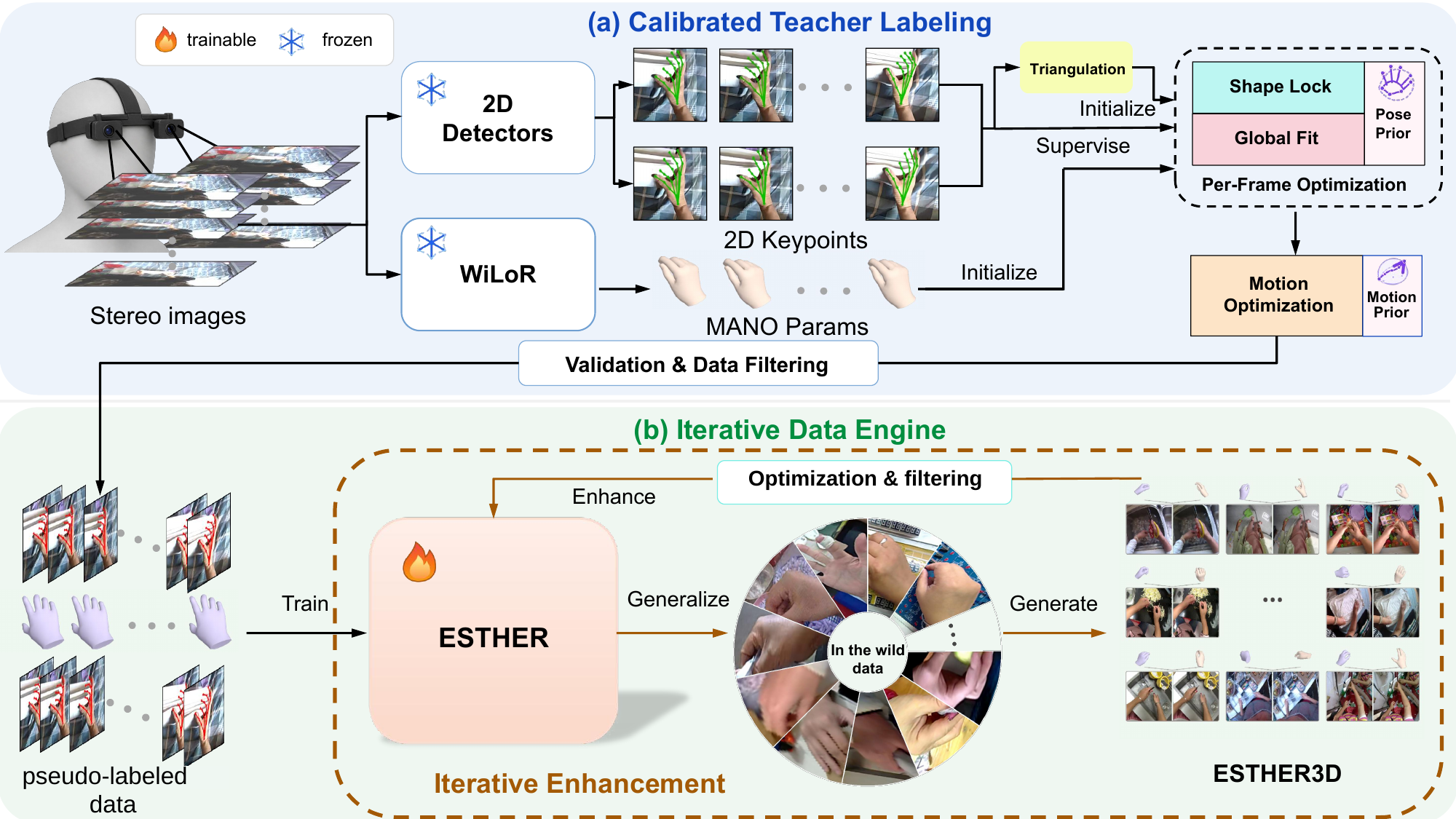}
\caption{\textbf{Supervision pipeline and data engine.} (a) The teacher
turns stereo 2D keypoints and WiLoR predictions from each view into metric pseudo labels via optimization in two stages and a reprojection/human filter. (b) These train
ESTHER, which labels in-the-wild stereo to build ESTHER3D.}
\label{fig:pipeline}
\end{figure*}

\subsection{Egocentric and Embodied Datasets}

Egocentric datasets are central to embodied perception:
EPIC-KITCHENS~\cite{epickitchens2018} and Ego4D~\cite{ego4d2022} cover
extended daily activities, Ego-Exo4D~\cite{egoexo4d2024} pairs egocentric
and external views, and Nymeria~\cite{nymeria2024} adds multimodal
wearable motion sensing.

Datasets of interactions between hands and objects such as FPHA~\cite{fpha2018},
H2O~\cite{h2o2021}, HOI4D~\cite{hoi4d2022}, OakInk~\cite{oakink2022},
DexYCB~\cite{dexycb2021}, ARCTIC~\cite{arctic2023},
AssemblyHands~\cite{assemblyhands2023}, and HOT3D~\cite{hot3d2024} provide
important supervision for hands and interaction. HOT3D is closest, but its
wearable stereo is monochrome and its metric labels come from motion
capture. ESTHER3D instead targets in-the-wild calibrated short-baseline RGB
stereo with metric MANO labels and train/test protocols.

\subsection{3D Hand Pose and Mesh Estimation}

Parametric hand reconstruction builds on MANO~\cite{mano2017}, supported
by datasets such as FreiHAND~\cite{freihand2019}, HO-3D~\cite{ho3d2019},
InterHand2.6M~\cite{interhand2020}, DexYCB~\cite{dexycb2021},
ObMan~\cite{obman2019}, and ARCTIC~\cite{arctic2023}. Monocular models
from I2L-MeshNet~\cite{i2lmeshnet2020}, Mesh
Graphormer~\cite{meshgraphormer2021}, MobRecon~\cite{mobrecon2022}, and
HandOccNet~\cite{handoccnet2022} to HaMeR~\cite{hamer2024} and
WiLoR~\cite{wilor2025} recover articulation well but must regress scale
and metric depth from learned priors alone.

On the temporal side, HTT~\cite{htt2023}, HandFormer~\cite{handformer2024},
and UmeTrack~\cite{umetrack2022} track hands in head-mounted settings, and
HaWoR~\cite{hawor2025} and Dyn-HaMR~\cite{dynhamr2025} recover motion in world coordinates from monocular video via monocular SLAM and offline optimization, with ambiguous scale. ESTHER instead obtains metric scale feedforward from
calibrated stereo and disentangles hand from head motion with conditioning
tokens and masked temporal training.

\subsection{Multiview, Stereo, and Depth-Based Reconstruction}

Multiview pose estimation has explored learnable triangulation, epipolar
attention, volumetric fusion, and cross-view
transformers~\cite{learnabletriangulation2019,epipolartransformers2020,
voxelpose2020,mvp2021,mvgformer2023}; POEM-v2~\cite{poemv2_2024} studies
multiview hand reconstruction with point embeddings. Closer to us, egocentric stereo has been
used for absolute hand pose~\cite{seo2021stereo} and stereo
keypoints~\cite{stereohand2022}, and event cameras for stereo hand
pose~\cite{egoevhandpose2025} and MANO mesh
reconstruction~\cite{eventegohands2025}. These estimate sparse keypoints,
use non-RGB sensing, or are single-frame; we target temporally consistent
metric MANO reconstruction from wearable RGB stereo, where neither
visibility in both views nor detection in each frame can be assumed.

\section{Method}

ESTHER is built together with its supervision pipeline
(Figure~\ref{fig:pipeline}). The calibrated teacher labels an initial
10 hours of in-the-wild stereo using 2D evidence and a DPoser-X
prior~\cite{dposerx2025}. ESTHER predictions then initialize teacher
optimization on more samples; filtered labels feed back into training
(Section~\ref{sec:training}). The final ESTHER supplies forward predictions
for deployment and testing; the separate mocap collection provides metric
evaluation.

\subsection{Calibrated Teacher Labeling Pipeline}

The teacher starts from 2D evidence, which is far easier to obtain reliably than metric 3D hands, and converts it into metric MANO pseudo labels via stereo calibration,
hand priors, and optimization. We design a simple head-mounted RGB stereo
rig (camera intrinsics and extrinsics in Appendix~F); the formulation is
not tied to it, and our experiments show the model adapts to different
capture setups with only minimal fine-tuning.

\begin{figure*}[t]
\centering
\captionsetup[subfigure]{font=footnotesize,skip=3pt,justification=centering}
\newlength{\archframeheight}\setlength{\archframeheight}{0.2835\textwidth}
\newlength{\archimageheight}\setlength{\archimageheight}{0.314\textwidth}
\newlength{\archstereowidth}\setlength{\archstereowidth}{0.7220642\archframeheight}
\newlength{\archmotionwidth}\setlength{\archmotionwidth}{0.8035264\archframeheight}
\begin{subfigure}[b]{0.555\textwidth}
\centering
\includegraphics[width=\linewidth]{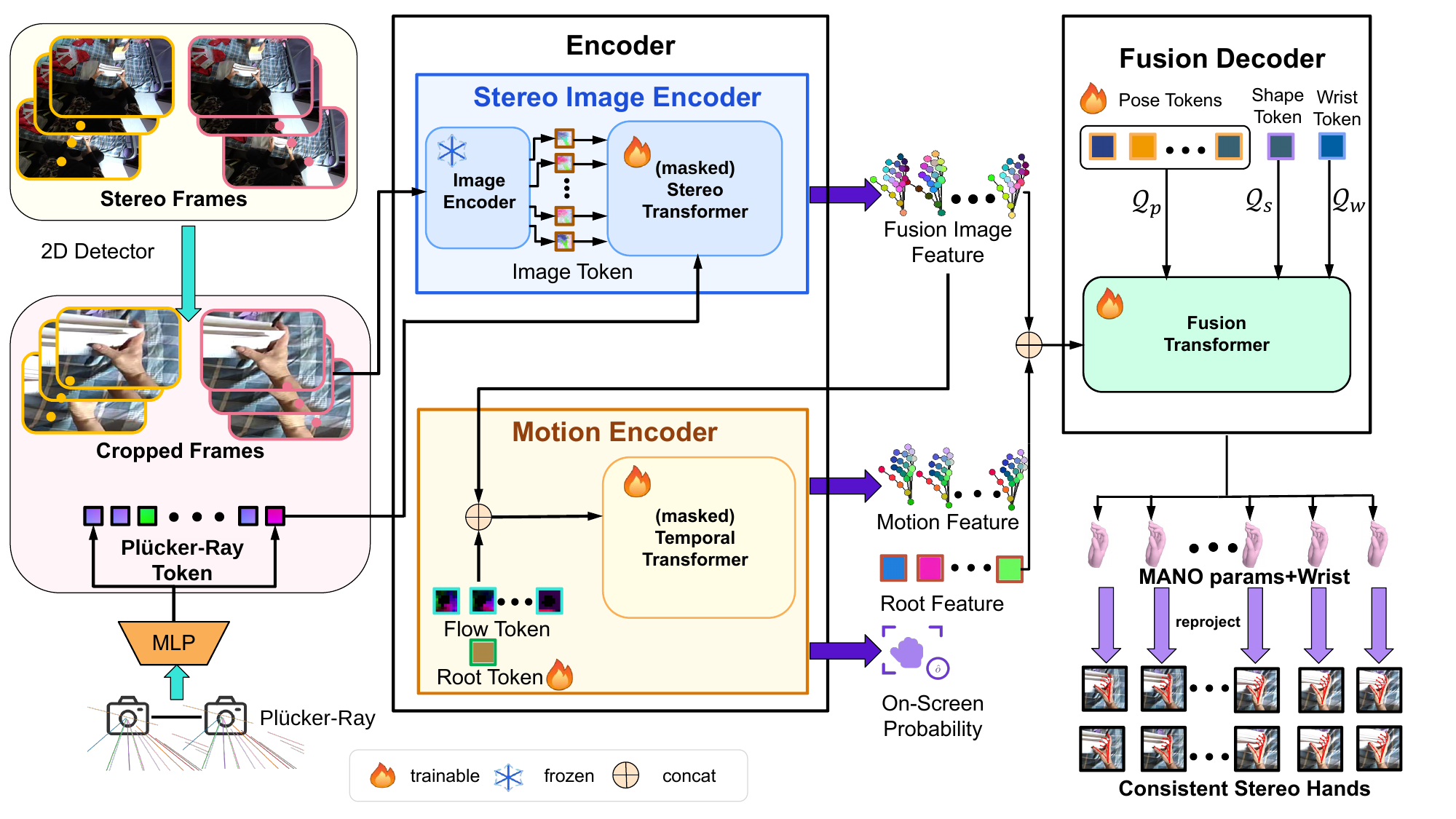}
\caption{ESTHER}
\label{fig:model_a}
\end{subfigure}\hfill
\begin{subfigure}[b]{0.205\textwidth}
\centering
\begin{minipage}[t][\archimageheight][t]{\linewidth}\centering
\includegraphics[height=\archframeheight,trim=0bp 44.8bp 0bp 3bp,clip]{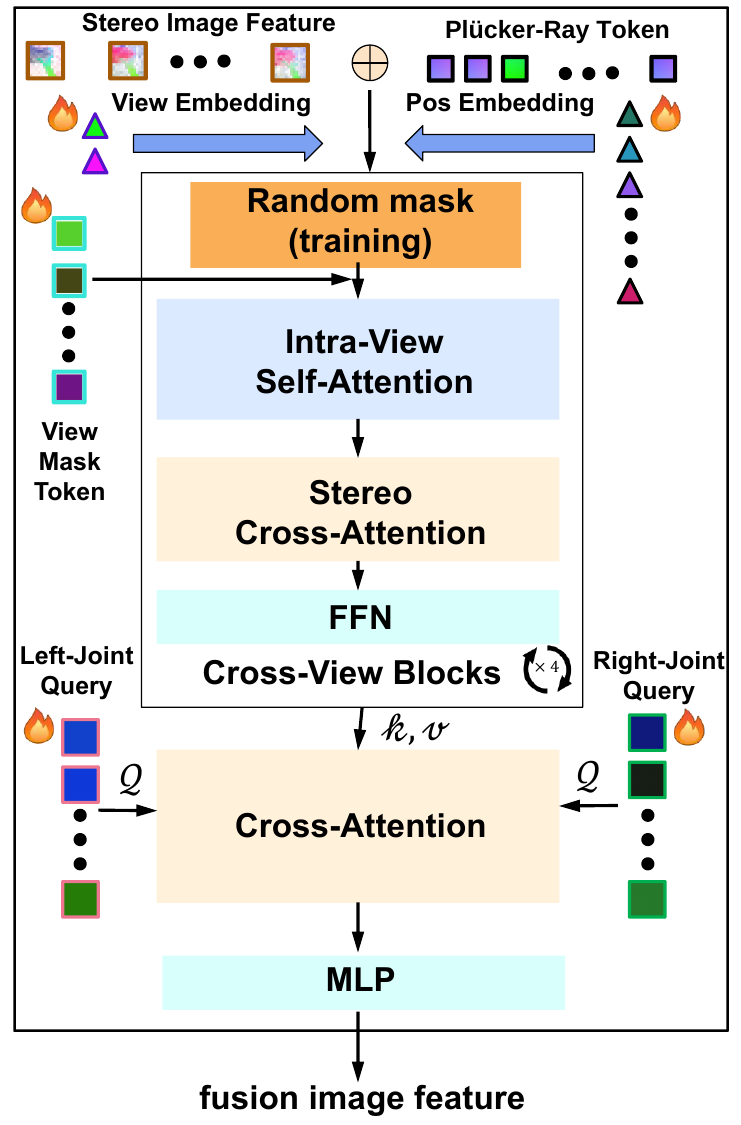}
\par\nointerlineskip
\includegraphics[width=\archstereowidth,trim=0bp 0bp 0bp 495.2bp,clip]{Figures/arch_stereo_vector.pdf}
\end{minipage}
\caption{Stereo Transformer}
\label{fig:model_b}
\end{subfigure}\hfill
\begin{subfigure}[b]{0.230\textwidth}
\centering
\begin{minipage}[t][\archimageheight][t]{\linewidth}\centering
\includegraphics[height=\archframeheight,trim=0bp 48.8bp 0bp 6bp,clip]{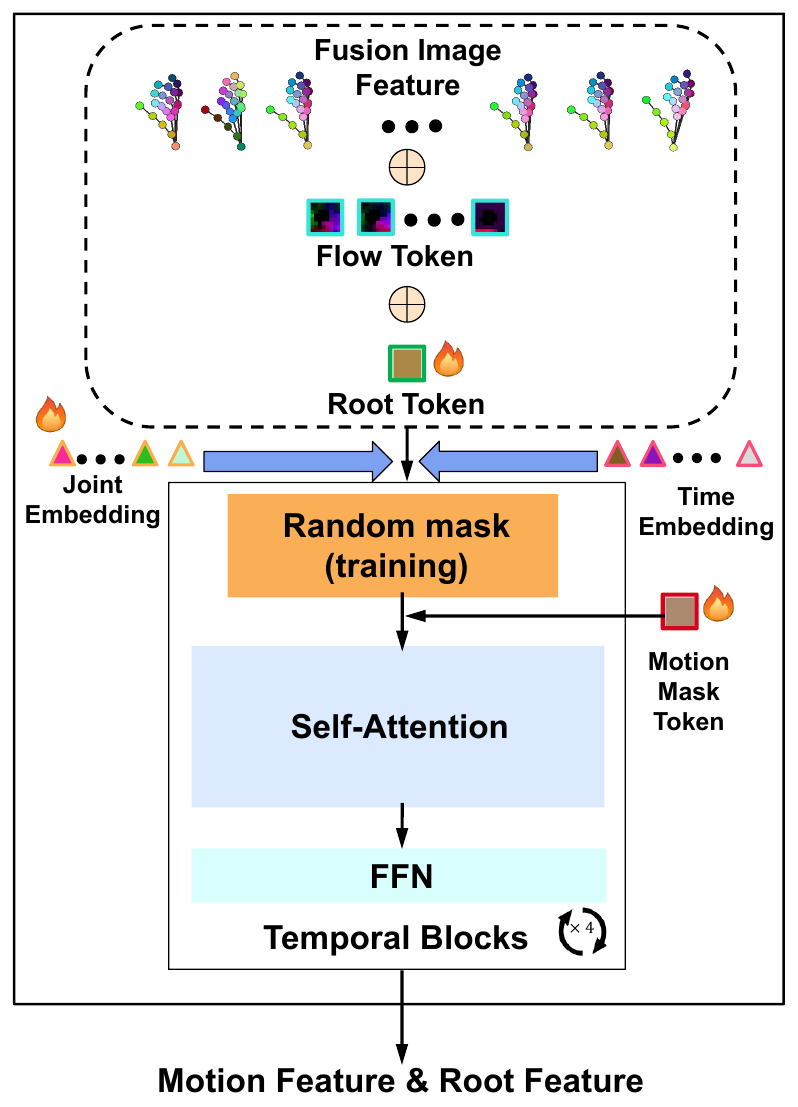}
\par\nointerlineskip
\includegraphics[width=\archmotionwidth,trim=0bp 0bp 0bp 482.4bp,clip]{Figures/arch_motion_vector.pdf}
\end{minipage}
\caption{Temporal Transformer}
\label{fig:model_c}
\end{subfigure}
\caption{\textbf{ESTHER.} (a) The stereo image encoder and masked temporal
motion encoder feed a query-gated fusion decoder that outputs MANO with a
metric wrist. (b, c) Stereo and temporal transformer blocks.}
\label{fig:model}
\end{figure*}

An ensemble of 2D detectors
(YOLO11-pose~\cite{yolo11_ultralytics} and ViTPose~\cite{vitpose2022})
provides keypoints on both views, triangulated through the calibrated rig
and filtered by depth range and plausibility of bone lengths. Optimization is initialized from the triangulated wrist and
monocular MANO estimates from each view from WiLoR~\cite{wilor2025}, and proceeds
in two stages under the DPoser-X pose prior.
Shape lock: on the 64 most confident frames, shape, pose, and
wrist are jointly optimized against the filtered stereo observations, and
the shapes estimated for individual frames are averaged into one fixed $\bar{\boldsymbol\beta}$
per hand. Pose and motion optimization: with the shape locked, pose
and wrist are reoptimized per frame, then refined by 16-frame
sliding-window motion optimization with velocity limits and an
motion prior pretrained on AMASS~\cite{amass2019,humor2021,hmp2024}. The
detection filtering thresholds, initialization, full fitting objective, and
stage details are given in Appendix~A.

\subsection{Ground Truth Validation}

\begin{figure}[t]
\centering
\includegraphics[width=0.90\columnwidth]{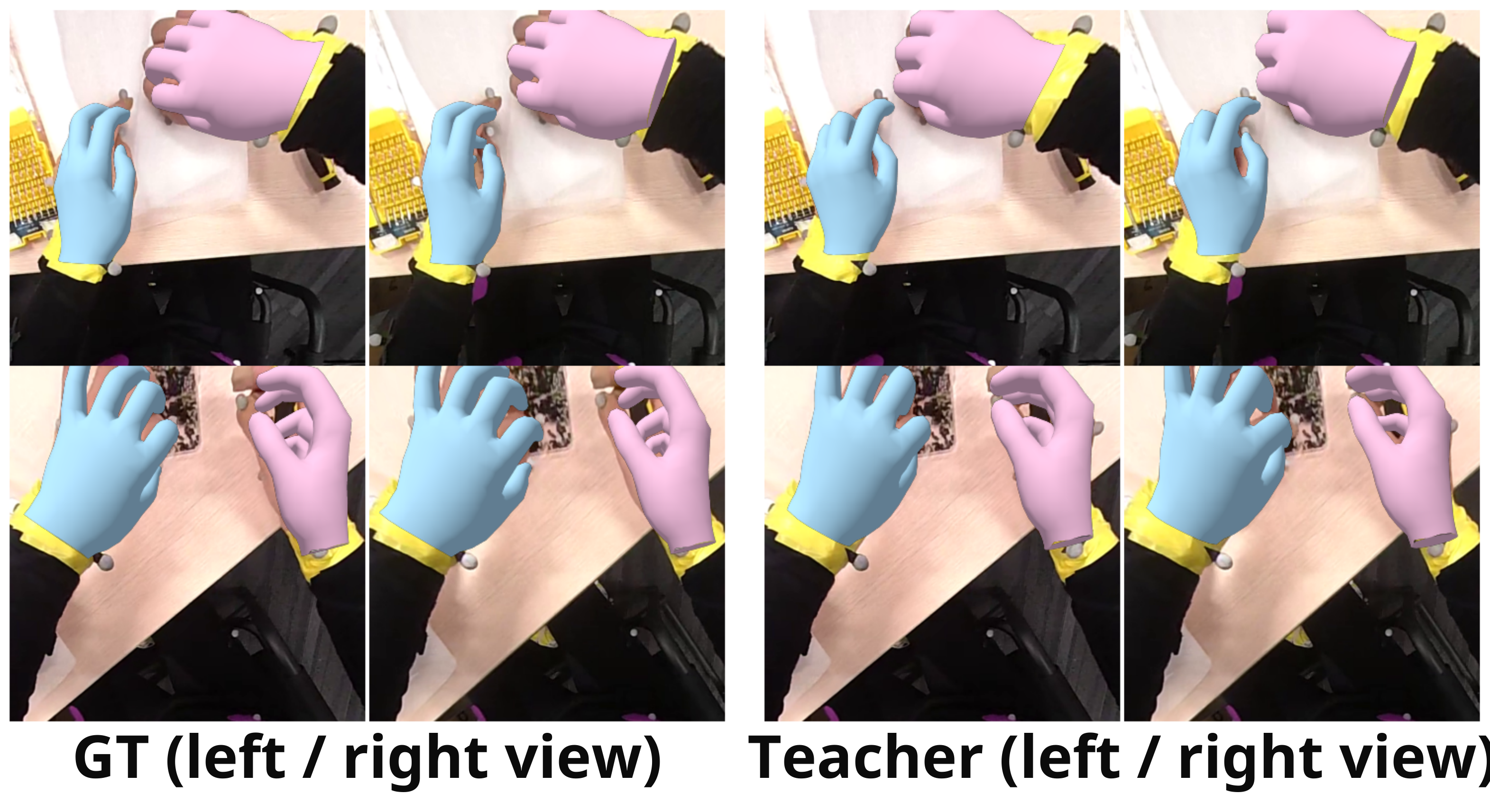}
\caption{Teacher pseudo labels vs.\ motion capture ground truth as MANO
meshes projected into both stereo views (two ESTHER3D test scenes; the GT
mesh is MANO fitted to the mocap joints).}
\label{fig:teacher_quad}
\end{figure}

Before scaling up, we verify that the automatic labels can serve as
supervision on the full 10-hour motion capture test set (Appendix~H). The teacher runs
on it without access to the labels and reaches 15.9\,mm MPJPE, 9.3\,mm
PA-MPJPE, and 19.1\,mm wrist error zero-shot (Figure~\ref{fig:teacher_quad}).
This approaches the roughly 10\,mm uncertainty in solving glove measurements of the ground
truth itself and is on par with the quality of annotations obtained by fitting of
widely used hand benchmarks~\cite{freihand2019,ho3d2019}, meeting the
accuracy required of pseudo labels; the full measurement chain is in
Appendix~I. The validation also exposes failure modes (2D misdetections under
heavy occlusion), handled by confidence thresholds and manual quality
control.

\subsection{In-the-Wild Data and Filtering}

The validated teacher is applied to a subset of the raw in-the-wild
collection, 10 hours drawn evenly from the six categories (collection
statistics in the ESTHER3D section). Labels pass a filter with three stages: a coarse geometric
prescreen on bone length and depth range between hand and camera, an automatic 2D check
that reprojects each fitted hand into both views and discards hand frames
disagreeing with the detections, and human review for the residual failures
these cannot catch (wrong hand identity, implausible articulation).
Qualified data trains ESTHER; its predictions initialize teacher refinement
and the same filtering on more data (ESTHER3D Benchmark section).

\subsection{Model Overview}

ESTHER starts from the standard pipeline of hand detection and cropping~\cite{mobrecon2022,hamer2024,wilor2025}: a detector produces
left/right crops with their virtual crop camera calibration, and the
network consumes only these. Training uses right hands: left hands are mirrored with views swapped and predictions unmirrored, following WiLoR~\cite{wilor2025},
halving the pose space to learn.

The model consists of two encoders and a decoder
(Figure~\ref{fig:model}):
\begin{equation}
\begin{aligned}
F&=E_{\mathrm{img}}(I^l,I^r,\boldsymbol{\ell}^l,\boldsymbol{\ell}^r),\\
Z,\mathbf{z}^{root},\hat{o}&=E_{\mathrm{mot}}(F,\mathbf{f},\mathbf{r}),\\
\hat{\Theta}&=D_{\mathrm{fuse}}(F,Z,\mathbf{z}^{root}),
\end{aligned}
\end{equation}
where, per hand, $\hat{\Theta}=(\hat{\boldsymbol{\theta}},\hat{\boldsymbol{\beta}},
\hat{\mathbf{t}})$ collects the MANO pose ($16\times3$ axis-angle), the
10-dimensional MANO shape, and the 3-dimensional global wrist translation.
$I^l,I^r$ are cropped stereo images and $\boldsymbol{\ell}^l,
\boldsymbol{\ell}^r$ the Plucker rays of their patches, computed from the
crop camera calibration.
At each frame, the stereo image encoder $E_{\mathrm{img}}$ fuses the two views
into 21 joint tokens $F$. The temporal motion encoder
$E_{\mathrm{mot}}$, conditioned on a camera motion flow token $\mathbf{f}$
and a learned root token $\mathbf{r}$, outputs a motion feature $Z$, a root
feature $\mathbf{z}^{root}$, and hand visibility probabilities $\hat{o}$
that gates, at deployment, whether the decoded hand is emitted at all. The fusion decoder reads all sources with
structured queries.
During training the image encoder additionally emits 2D keypoints,
visibility, and triangulated anchors as guidance, and the motion encoder is
trained on randomly masked sequences;
heads and losses are detailed in Appendix~B.

\subsection{Stereo Image Encoder}

The stereo image encoder pairs a frozen DINOv3-L image
encoder~\cite{dinov3_2025} with a stereo transformer
(Figure~\ref{fig:model_b}). It encodes the 16-frame clip together but injects
no temporal information, fusing the two views purely through complementary
image evidence; all temporal modeling is left to the motion encoder. Patch
tokens of each view are augmented with embeddings of their
calibrated Plucker rays, situating independently detected crops in the
common metric geometry, and pass through alternating intra-view and
cross-view attention, so spatial context within each view and cross-view stereo
matching refine each other layer by layer; 21 joint queries per view then
read from these fused tokens and are aggregated into the shared feature $F$. During training one entire view is randomly blanked and replaced by a
learned view mask token, so the fusion explicitly survives a missing camera.
Lightweight heads (heatmaps, 2D keypoints, visibility, depth bins, a
differentiable triangulation) guide the encoder toward explicit stereo
geometry. The DINOv3-L backbone stays frozen throughout; only the
stereo transformer and its heads are trained, so the encoder inherits strong
pretrained features while keeping the trainable parameter count small
(51M of 355M total).

\subsection{Masked Temporal Motion Encoder}

The motion encoder provides the temporal reasoning that egocentric capture
demands. It feeds three inputs directly to a temporal transformer
(Figure~\ref{fig:model_c}): the fusion image feature $F$ (with joint and
time embeddings), the camera motion flow token $\mathbf{f}$, and a learned
root token $\mathbf{r}$ that aggregates global hand motion into the root
feature $\mathbf{z}^{root}$.

\textbf{Detection dropout and occlusion.}
Tokens are randomly masked per joint, per frame, and over contiguous spans
(anchor frames kept), each masked slot filled by a learned motion mask token
(distinct from the stereo view mask token), and auxiliary heads reconstruct
joints, pose, and root velocity there: the encoder learns to inpaint exactly
the corruption of in-the-wild capture. A head predicts framewise hand visibility, using blank frames as negatives.

\textbf{Coupling of hand and head motion.}
We compute dense classical optical flow online from consecutive RGB hand
crops, including both hand and background pixels. The previous crop is
warped into the current crop coordinates using their calibrated intrinsics
to compensate changes in crop position and scale. A fast estimator that
requires no training computes flow on this compensated pair. Median pooling
on an $8\times8$ grid gives 128 flow values; normalization by the 224-pixel
crop side and concatenation with 64 validity flags form the 192-dimensional
flow token (Appendix~D). This unmasked input carries both camera and hand
motion, providing context when image tokens are masked.

\subsection{Query-Gated Fusion Decoder}

The fusion decoder lets pose, shape, and wrist read different evidence.
As in Figure~\ref{fig:model_a}, $\mathbf Q_p$ contains 16 pose queries along
the kinematic tree, while $\mathbf Q_s$ and $\mathbf Q_w$ are the shape and
wrist queries. Each layer applies self-attention among these 18 queries,
yielding state $\mathbf h_i$ for query $i$, then reads
$S_{\mathrm{img}}=F$, $S_{\mathrm{mot}}=Z$, and
$S_{\mathrm{root}}=\mathbf z^{root}$,
$k\in\{\mathrm{img},\mathrm{mot},\mathrm{root}\}$, with a softmax gate $\mathbf{g}_i$ for each token:
\begin{equation}
\mathbf{q}^{+}_i=\operatorname{FFN}\Big(\mathbf{h}_i
+\sum_{k}g_i^k\operatorname{CrossAttn}(\mathbf{h}_i,S_k)\Big),
\end{equation}
where $\mathbf{q}^{+}_i$ is the updated token. The gate is predicted from
the token state, pooled source summaries, and evidence statistics such as
visibility and triangulation spread (Appendix~E): the pose tokens can emphasize
the image and motion features while the wrist token leans on the root
feature when image evidence is ambiguous.

Pose is decoded from $\mathbf Q_p$ along the kinematic tree, conditioning
each child rotation on its parent. After pooling over time, $\mathbf Q_s$ yields
one shape per sequence. The final wrist query $\mathbf Q_w^{\mathrm{out}}$
directly predicts metric wrist translation:
\begin{equation}
\hat{\mathbf{t}}=\operatorname{MLP}(\mathbf Q_w^{\mathrm{out}}) .
\end{equation}
MLP denotes a multilayer perceptron. We deliberately do not anchor the wrist to the triangulated point: an
anchored wrist improves MPJPE with both views present in the ablation but requires a
reliable stereo anchor. Triangulation
instead contributes through its guidance loss and the gating statistics,
and the decoded wrist survives monocular fallback. A MANO
forward kinematics layer maps $\hat{\Theta}$ to joints and vertices.

\subsection{Training and Supervision}
\label{sec:training}

\textbf{Iterative training flywheel.}
To limit labeling cost, the teacher optimizes an initial 10 hours of
in-the-wild video for approximately 2000 steps; filtered outputs train the
first ESTHER model. On additional samples, ESTHER forward inference
initializes MANO parameters, which the teacher refines with 200 steps of
the same optimization. The same filtering
(Appendix~A) selects teacher outputs to retrain
ESTHER. We repeat this cycle. Deployment and testing use forward inference
of the final ESTHER model.

We pretrain the stereo image encoder with geometric guidance, then jointly
train the full model with forward kinematics losses and intermediate encoder
supervision. The image encoder retains 2D and triangulation objectives;
the motion encoder reconstructs canonical joints, pose, and root velocity at
masked positions and predicts hand visibility, learning temporal structure
and inpainting (Appendix~B). Losses exclude invalid targets. Motion token
masking and image augmentation (noise, erasing, and blanking views and
frames, with blank frames as visibility negatives) simulate corruption
in the wild. The overall objective is
\begin{equation}
\mathcal{L}
=\mathcal{L}_{\mathrm{FK}}
+\mathcal{L}_{\mathrm{img}}
+0.5\mathcal{L}_{\mathrm{mot}}
+0.5\mathcal{L}_{\mathrm{on}},
\end{equation}
where $\mathcal{L}_{\mathrm{FK}}$ gathers joint, mesh, pose, shape, and
wrist terms, $\mathcal{L}_{\mathrm{img}}$ the image guidance,
$\mathcal{L}_{\mathrm{mot}}$ the masked reconstruction, and
$\mathcal{L}_{\mathrm{on}}$ the on-screen classification (all terms in
Appendix~B); no oracle signal is provided at test time. The full model trains
in about two days on four RTX~4090 GPUs.

\section{ESTHER3D Benchmark}
\begin{figure}[t]
\centering
\includegraphics[width=0.90\columnwidth]{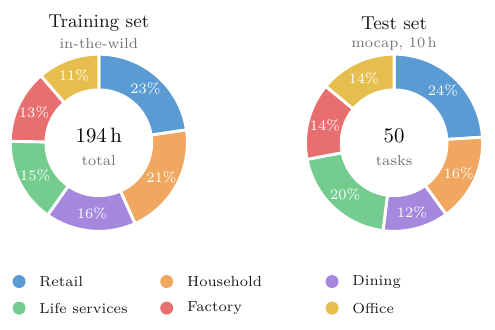}
\caption{ESTHER3D across the six industries. Left: in-the-wild training
hours (194\,h). Right: motion capture test tasks (50 tasks, $\approx$10\,h).}
\label{fig:hours}
\end{figure}

ESTHER3D consists of two parts: a large in-the-wild training set
carrying metric pseudo labels from ESTHER, and a motion capture
test set with true metric ground truth. Every frame carries MANO pose and shape for each hand, absolute metric wrist, calibrated camera parameters, 2D
keypoint projections in both views, and hand bounding boxes. The training
set comes from
a distributed program that collects raw in-the-wild stereo video with our
rig across six industries, yielding about 194 hours from 121 collectors in 308 real venues, 44 scene categories, and roughly 800 manipulation tasks
(Figure~\ref{fig:hours}; breakdown in Appendix~G). The collection covers common conditions for wearable systems: hands leaving the field of view, object interactions,
self-occlusion, rapid head motion, and asymmetric visibility
(Figure~\ref{fig:teaser}). The test set is captured in a motion capture studio mirroring the six
industries (about 10 hours, 6 subjects, 50 scripted tasks spanning 14
hand action types; Appendix~H).

\subsection{Labeling with ESTHER and Iterative Enhancement}

ESTHER labels the raw collection at scale, in the spirit of depth data
engines~\cite{depthanything2024,depthanythingv2_2024} but grounded by
stereo geometry and motion capture validation: each prediction is refined
by 200 steps of teacher optimization under the DPoser-X prior and the 2D
evidence (2D keypoints in both views), then passes the same
reprojection and human review filter as the teacher. Surviving
labels enter the benchmark and feed back into training through the
flywheel in Section~\ref{sec:training}.

\textbf{Label quality.}
The in-the-wild labels have a median reprojection residual of 5.3\,px,
corresponding to a median in-plane discrepancy of 2.8\,mm. With the
6.2\,cm stereo baseline, the propagation analysis in Appendix~N indicates
about 21\,mm depth sensitivity under independent view errors. These are
agreement and sensitivity measures, not bounds on true 3D label error.
Metric accuracy is assessed against motion capture: the teacher reaches
15.9\,mm MPJPE and 19.1\,mm wrist error.

\section{Experiments}

\begin{table}[t]
\centering
\caption{\textbf{Main comparison.} Errors in mm, jitter in m/s$^2$.
ESTHER3D test = real mocap GT, zero-shot (z.s.) for all methods; HOT3D
reported zero-shot and after fine-tuning (f.t.) on 10\% of HOT3D. Best per
block in bold; $\downarrow$ means lower is better.}
\label{tab:main}
\renewcommand{\arraystretch}{0.85}%
\resizebox{\columnwidth}{!}{%
\begin{tabular}{@{}cl cccc@{}}
\toprule
& Method & \makebox[4.8em]{MPJPE $\downarrow$} & PA-MPJPE $\downarrow$ & Wrist $\downarrow$ & Jitter $\downarrow$ \\
\midrule
& Hiera-FTL~\cite{hiera2023,hot3dchallenge1st2024} & 43.4 & 10.0 & 277 & 79.5 \\
& POEM-v2~\cite{poemv2_2024} & 19.4 & 8.9 & 22.5 & 8.5 \\
\raisebox{0pt}[0pt][0pt]{\rotatebox[origin=c]{90}{ESTHER3D}} & MVGFormer~\cite{mvgformer2023} & 19.2 & 8.9 & 21.0 & 8.7 \\
& Epipolar Trans.~\cite{epipolartransformers2020} & 19.0 & 9.0 & \textbf{20.9} & 13.3 \\
& \cellcolor{gray!12}ESTHER (ours) & \cellcolor{gray!12}\textbf{17.7} & \cellcolor{gray!12}\textbf{8.6} & \cellcolor{gray!12}21.3 & \cellcolor{gray!12}\textbf{3.3} \\
\midrule
& Hiera-FTL~\cite{hiera2023,hot3dchallenge1st2024} & 103.2 & 18.9 & 324.3 & 99.6 \\
& MVGFormer~\cite{mvgformer2023} & 52.2 & 18.7 & 71.9 & 31.5 \\
\raisebox{0pt}[0pt][0pt]{\rotatebox[origin=c]{90}{HOT3D z.s.}} & POEM-v2~\cite{poemv2_2024} & 49.7 & \textbf{16.7} & 66.4 & 33.4 \\
& Epipolar Trans.~\cite{epipolartransformers2020} & 48.3 & 19.5 & 83.8 & 51.1 \\
& \cellcolor{gray!12}ESTHER (ours) & \cellcolor{gray!12}\textbf{30.5} & \cellcolor{gray!12}17.3 & \cellcolor{gray!12}\textbf{54.2} & \cellcolor{gray!12}\textbf{20.9} \\
\midrule
& Hiera-FTL~\cite{hiera2023,hot3dchallenge1st2024} & 45.3 & 13.6 & 46.0 & 52.8 \\
& POEM-v2~\cite{poemv2_2024} & 35.5 & 9.5 & 25.9 & 23.7 \\
\raisebox{0pt}[0pt][0pt]{\rotatebox[origin=c]{90}{HOT3D f.t.}} & Epipolar Trans.~\cite{epipolartransformers2020} & 35.1 & 9.6 & 45.4 & 55.9 \\
& MVGFormer~\cite{mvgformer2023} & 34.9 & 9.2 & 24.7 & 23.2 \\
& \cellcolor{gray!12}ESTHER (ours) & \cellcolor{gray!12}\textbf{29.5} & \cellcolor{gray!12}\textbf{8.0} & \cellcolor{gray!12}\textbf{20.8} & \cellcolor{gray!12}\textbf{17.1} \\
\bottomrule
\end{tabular}
}
\end{table}

\begin{figure*}[t]
\centering
\includegraphics[width=0.80\textwidth]{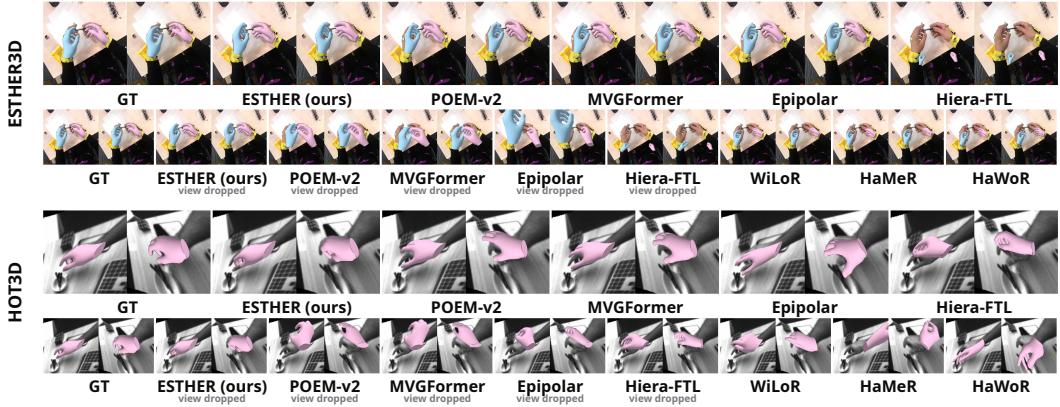}
\caption{Qualitative results on ESTHER3D and HOT3D: MANO meshes projected
into both stereo views (pink right, blue left hand). Per block: top row full
stereo, bottom row monocular.}
\label{fig:mocap_arch}
\end{figure*}

The main text reports the controlled architecture comparison, the
robustness studies, and the component ablation; remaining details are in
Appendices~A--O.

\begin{table}[t]
\centering
\caption{\textbf{Robustness to missing views.} One stereo view is dropped for
methods trained with stereo (ESTHER falls back via its view mask token);
monocular methods take their usual single view. Metrics as in
Table~\ref{tab:main}.}
\label{tab:viewdrop}
\renewcommand{\arraystretch}{0.85}%
\resizebox{\columnwidth}{!}{%
\begin{tabular}{@{}cl cccc@{}}
\toprule
& Method & \makebox[4.8em]{MPJPE $\downarrow$} & PA-MPJPE $\downarrow$ & Wrist $\downarrow$ & Jitter $\downarrow$ \\
\midrule
& Hiera-FTL~\cite{hiera2023,hot3dchallenge1st2024} & 65.2 & 9.9 & 217 & 55.2 \\
& HaWoR~\cite{hawor2025} (mono, temporal) & 43.8 & 11.2 & 97.3 & 4.8 \\
& HaMeR~\cite{hamer2024} (mono) & 36.3 & 11.8 & 100 & 10.5 \\
\raisebox{0pt}[0pt][0pt]{\rotatebox[origin=c]{90}{ESTHER3D}} & WiLoR~\cite{wilor2025} (mono) & 30.2 & 11.1 & 101.6 & 9.5 \\
& POEM-v2~\cite{poemv2_2024} & 29.5 & 9.7 & 96.8 & 8.8 \\
& MVGFormer~\cite{mvgformer2023} & 28.9 & 9.8 & 97.1 & 9.3 \\
& Epipolar Trans.~\cite{epipolartransformers2020} & 27.0 & 9.5 & 171.7 & 10.5 \\
& \cellcolor{gray!12}ESTHER (ours) & \cellcolor{gray!12}\textbf{18.0} & \cellcolor{gray!12}\textbf{9.0} & \cellcolor{gray!12}\textbf{32.5} & \cellcolor{gray!12}\textbf{4.3} \\
\midrule
& HaWoR~\cite{hawor2025} (mono, temporal) & 71.8 & 29.3 & 214 & 26.9 \\
& HaMeR~\cite{hamer2024} (mono) & 69.5 & 17.4 & 126.8 & 96.4 \\
& Hiera-FTL~\cite{hiera2023,hot3dchallenge1st2024} & 52.4 & 18.7 & 115.6 & 53.3 \\
\raisebox{0pt}[0pt][0pt]{\rotatebox[origin=c]{90}{HOT3D f.t.}} & Epipolar Trans.~\cite{epipolartransformers2020} & 42.1 & 15.8 & 93.2 & 45.8 \\
& MVGFormer~\cite{mvgformer2023} & 39.9 & 14.6 & 90.0 & 23.3 \\
& POEM-v2~\cite{poemv2_2024} & 37.2 & 14.9 & 130.2 & 22.7 \\
& WiLoR~\cite{wilor2025} (mono) & 31.0 & 15.9 & 111.5 & 91.8 \\
& \cellcolor{gray!12}ESTHER (ours) & \cellcolor{gray!12}\textbf{30.2} & \cellcolor{gray!12}\textbf{12.1} & \cellcolor{gray!12}\textbf{33.9} & \cellcolor{gray!12}\textbf{18.4} \\
\bottomrule
\end{tabular}
}
\end{table}

\subsection{Protocol and Metrics}

All trainable methods share the frozen backbone, data, losses, and budget,
so the comparison isolates architecture.
Evaluation covers the ESTHER3D test set (zero-shot for every method) and
HOT3D~\cite{hot3d2024}, zero-shot and after fine-tuning on 10\% of its
data. MPJPE is mean joint position error relative to the wrist; PA-MPJPE adds
Procrustes alignment. MPJPE-absolute denotes error without wrist alignment.
We report MPJPE, PA-MPJPE, absolute wrist error (mm), and temporal jitter
(m/s$^2$; Appendix~L). Ablations use MPJPE on the test set.

\subsection{Architecture Comparison}

\textbf{Baselines.}
We compare three groups: (1) multiview architectures adapted to stereo for individual frames:
Epipolar Transformers~\cite{epipolartransformers2020},
MVGFormer~\cite{mvgformer2023}, POEM-v2~\cite{poemv2_2024}; (2) the HOT3D challenge winner, using Hiera~\cite{hiera2023} with FTL camera fusion~\cite{hot3dchallenge1st2024}; and (3) pretrained monocular models: WiLoR~\cite{wilor2025}, HaMeR~\cite{hamer2024}, and the
temporal HaWoR~\cite{hawor2025}.

\textbf{Monocular wrist.}
The monocular methods predict a weak-perspective camera on the normalized
crop, $x_{\mathrm{crop}}=s\,(X+t_x)$, which involves no camera intrinsics.
Because the MANO joints $X$ are in meters, the scale $s$ carries units of
normalized crop per meter: metric scale already enters through MANO the
moment $s$ is predicted, and the hand lives directly in MANO metric
coordinates. Writing the same point with the true pinhole,
$u=f\,(X+t_x)/t_z+c$ and $x_{\mathrm{crop}}=2(u-c_{\mathrm{box}})/b$ for a
crop box of side $b$ pixels, and matching the coefficient of $X$ gives
$s=2f/(t_z b)$, \ie, $t_z=2f/(s\,b)$, which is dimensionally meters. Recovering the wrist depth is thus a pure
unit conversion rather than an intrinsic assumption, so their wrists are
evaluated in the same metric camera frame as ESTHER's.

\textbf{ESTHER3D test set.}
Against real ground truth (zero-shot for every method;
Figure~\ref{fig:mocap_arch}), ESTHER leads where stereo and temporal fusion
matter: best pose (17.7\,mm MPJPE) and three to four times lower jitter
(3.3 versus 8.5--13.3\,m/s$^2$). Independent frame processing cannot model this temporal behavior. ESTHER's wrist error on clean frames matches the multiview methods (21.3 versus 20.9--22.5\,mm; absolute wrist prediction trades some accuracy on clean frames for robustness to missing views in
Table~\ref{tab:viewdrop}), while Hiera-FTL's absolute head without an anchor fails
outright (277\,mm wrist).

\textbf{External generalization.}
On HOT3D, ESTHER's zero-shot MPJPE (30.5\,mm) beats even the fine-tuned
baselines. After fine-tuning, ESTHER achieves the lowest MPJPE, PA-MPJPE,
wrist error, and jitter among compared methods. 

ESTHER is also stable across the HOT3D pinhole focal sweep (Appendix~M).

\subsection{Robustness}

\textbf{One stereo view missing.}
We blank one view at inference, simulating a hand leaving or becoming occluded in one camera.

With a camera gone, the stereo baselines collapse in wrist: their
triangulated anchor vanishes with the view. ESTHER degrades gracefully
rather than collapsing: relative pose is nearly unchanged (17.7 $\rightarrow$
18.0\,mm MPJPE, PA-MPJPE 8.6 $\rightarrow$ 9.0\,mm), while absolute wrist error increases with the loss of stereo depth cues, rising from 21.3 to 32.5\,mm on ESTHER3D. This
is below the baselines' wrist errors of 96.8--217\,mm; training with blanked views and the view mask token let the model
fall back to a single view without the wrist collapse that undoes the
triangulation-anchored baselines. Even on a single grayscale view ESTHER
surpasses the 1.3B-parameter video diffusion model
ViDiHand~\cite{vidihand2026} at 355M total parameters (51M trainable atop a
frozen DINOv3-L backbone). This robustness comes from the stereo training
regime and the view mask token together: stereo supervision instills a metric
depth prior, and under it the token learns to complete depth from one view, as the brain does from a single eye, using scene context
(objects, arm) to stay stable and generalizable when a view is lost
(Sec.~\ref{sec:mono}). On HOT3D, inference from one view gives
30.2\,mm MPJPE, 12.1\,mm PA-MPJPE, 33.9\,mm wrist error, and
18.4\,m/s$^2$ jitter, the lowest values among the compared methods.

\begin{figure*}[t]
\centering
\begin{minipage}[b]{0.38\textwidth}
\centering
\includegraphics[width=\linewidth]{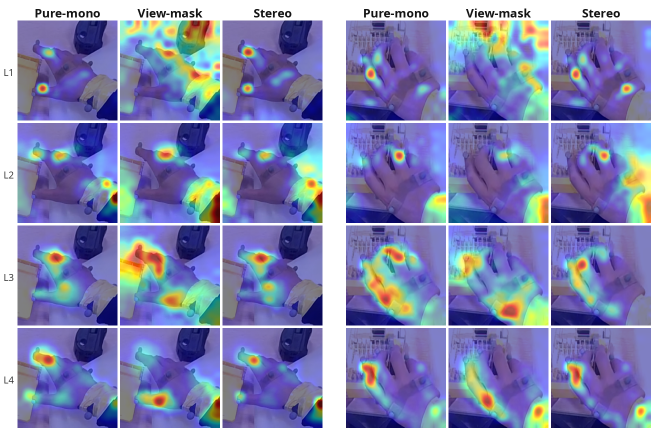}
\caption{Cross-attention into the kept view (two held-out scenes, four
transformer blocks L1--L4). Pure monocular and stereo attend almost
identically to the hand; the view mask placeholder instead spreads onto
surrounding objects and the arm, using scene context as a depth reference.}
\label{fig:monoattn}
\end{minipage}\hspace{0.06\textwidth}
\begin{minipage}[b]{0.44\textwidth}
\centering
\nextfloat
\captionsetup[subfigure]{font=footnotesize,skip=1pt,justification=centering}
\begin{subfigure}[b]{0.333\linewidth}\centering
\includegraphics[width=\linewidth]{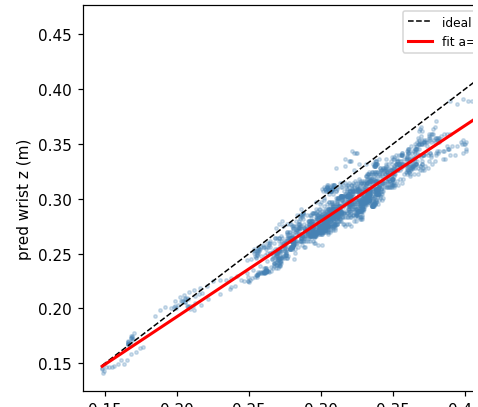}
\caption{stereo}\label{fig:monoscatter_a}
\end{subfigure}\hfill
\begin{subfigure}[b]{0.333\linewidth}\centering
\includegraphics[width=\linewidth]{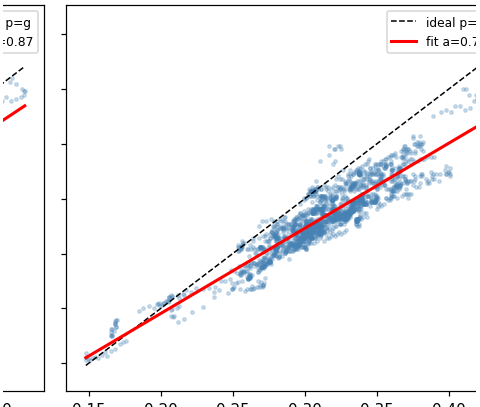}
\caption{view mask token}\label{fig:monoscatter_b}
\end{subfigure}\hfill
\begin{subfigure}[b]{0.333\linewidth}\centering
\includegraphics[width=\linewidth]{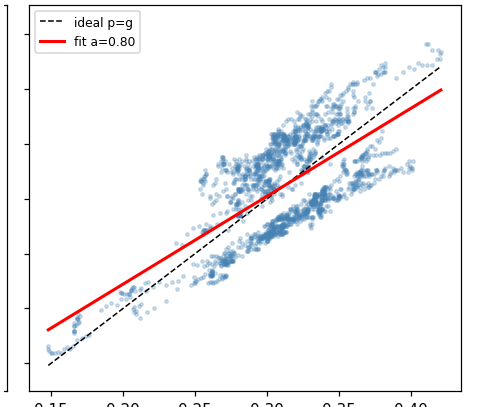}
\caption{pure-mono}\label{fig:monoscatter_c}
\end{subfigure}
\caption{Held-out wrist depth prediction vs.\ ground truth for (a) stereo,
(b) the view mask model with one view dropped, and (c) a separately trained
pure monocular model. (a) and (b) track true depth (correlation
$0.96$/$0.92$, random scatter $8$/$11$\,mm); (c) collapses (correlation
$0.69$, random scatter $32$\,mm).}
\label{fig:monoscatter}
\end{minipage}
\end{figure*}

\begin{table}[t]
\centering
\caption{Robustness to dropped frames as $N$ of 16 frames are blanked
and masked at inference, on the ESTHER3D test set and HOT3D. Each cell:
MPJPE / PA-MPJPE / wrist (mm) / jitter (m/s$^2$).}
\label{tab:frame_drop}
\resizebox{0.9\columnwidth}{!}{%
\begin{tabular}{@{}ccc@{}}
\toprule
$N$ dropped / 16 & ESTHER3D & HOT3D \\
\midrule
0 & 17.7 / 8.6 / 21.3 / 3.3 & 29.5 / 8.0 / 20.8 / 17.1 \\
2 & 17.8 / 8.6 / 21.3 / 3.9 & 29.7 / 8.2 / 20.9 / 18.2 \\
4 & 17.8 / 8.6 / 21.4 / 4.7 & 29.8 / 8.1 / 20.9 / 19.5 \\
6 & 17.8 / 8.6 / 21.5 / 7.7 & 29.8 / 8.1 / 21.2 / 23.8 \\
8 & 17.8 / 8.6 / 21.5 / 5.4 & 30.0 / 8.2 / 21.0 / 20.6 \\
\bottomrule
\end{tabular}
}
\end{table}

\textbf{Dropped frames.}
We drop $N$ random frames from each 16-frame clip (images blanked and
tokens masked), simulating the detection dropout that is routine in
egocentric capture, where fast motion and occlusion break detection in individual frames. Dropping up to half the
frames (8 of 16) leaves MPJPE, PA-MPJPE, and wrist essentially unchanged
on both datasets (Table~\ref{tab:frame_drop}). From $N=0$ to $N=8$, ESTHER3D MPJPE shifts by 0.1\,mm and wrist error by 0.2\,mm. Jitter increases overall but not monotonically with $N$, reflecting the extra frames interpolated between surviving
observations. This graceful degradation comes from the objective for masked frames, which teaches the motion encoder to inpaint missing frames from
temporal context. Appendix~J reports bootstrap confidence intervals for
these gaps and Appendix~K a study of failure modes under lighting, motion blur,
sensor noise, and skin tone shifts.

\subsection{Why the Monocular Fallback Stays Reliable}
\label{sec:mono}

Table~\ref{tab:viewdrop} shows that dropping a stereo view costs ESTHER only
a graceful wrist increase, not the collapse the baselines based on anchors
suffer. We probe why, comparing three configurations on the held-out
motion capture set: stereo (both views), view mask (our model
with one view dropped, its learned placeholder token standing in for the
missing view), and pure monocular, a model trained from scratch
with the second view always removed (a separate checkpoint, not our model run
monocularly).

\textbf{Attention (Figure~\ref{fig:monoattn}).} We visualize, across the four
stereo transformer blocks, the cross-attention that flows into the kept view.
The pure monocular model and the stereo model attend almost identically: both concentrate on the hand itself. The view mask model is visibly
different: its placeholder token queries beyond the hand, onto
surrounding objects, the workspace, and the arm. Trained under
stereo guidance, the token has learned to use scene context as a depth
reference, analogous to the monocular depth prior used by the human visual system with one eye. This supports feedforward inference without disparity.

\textbf{Depth scatter across frames (Figure~\ref{fig:monoscatter}).} We fit
$\hat{z}=a\,z_{\mathrm{gt}}+b$ to the predicted wrist depth on held-out data
and decompose the error into a systematic (linearly correctable) and a
random (irreducible) part. Stereo tracks true depth tightly
(correlation $0.96$, random scatter $8$\,mm). The view mask model stays close
(correlation $0.92$, random scatter $11$\,mm): it still knows how depth
varies frame to frame, and its residual error is mostly a systematic scale
compression that a single linear term removes. The pure monocular model,
despite being trained end-to-end for the monocular setting, collapses on
held-out data: correlation drops to $0.69$ and random scatter triples to
$32$\,mm, i.e.\ it regresses toward a prior mean plus noise. The view mask
token thus keeps depth estimates generalizable with low variance where
a dedicated monocular model overfits its training cues, which is exactly why
ESTHER degrades gracefully under a missing view.

\subsection{Ablation}

\begin{table}[t]
\centering
\caption{Component ablation on the ESTHER3D test set (MPJPE, mm).
$\Delta$ is the change from the full model; negative values indicate
improvement.}
\label{tab:ablation}
\resizebox{0.82\columnwidth}{!}{%
\begin{tabular}{@{}lcc@{}}
\toprule
Variant & MPJPE $\downarrow$ & $\Delta$ vs full \\
\midrule
Full ESTHER & 17.7 & -- \\
\midrule
w/o Plucker ray embedding & 27.7 & +10.0 \\
w/o cross-view attention & 26.4 & +8.7 \\
w/o all guidance, masking, augmentation & 20.9 & +3.2 \\
w/o flow token & 20.5 & +2.8 \\
w/o 2D / heatmap guidance & 19.2 & +1.5 \\
w/o image augmentation & 18.5 & +0.8 \\
uniform gate (concatenation) & 18.0 & +0.3 \\
w/o triangulation guidance & 17.8 & +0.1 \\
triangulation-anchored wrist & 15.5 & $-$2.2 \\
\bottomrule
\end{tabular}
}
\end{table}

Table~\ref{tab:ablation} identifies stereo fusion as the core contributor,
supported by complementary motion cues and auxiliary guidance. Geometric
wrist anchoring also benefits articulated pose: since MPJPE is
wrist-relative, the improvement reflects joint configuration and cannot
arise solely from more accurate absolute wrist localization.

\section{Conclusion}

We presented ESTHER, an end-to-end model for metric 3D hand reconstruction
from in-the-wild egocentric stereo, with ESTHER3D, a benchmark labeled by the model and validated by motion capture. ESTHER reaches state-of-the-art accuracy
with superior temporal smoothness and generalization, and a binding effect between size and depth keeps its metric scale under a lost view, dropped frames, or
monocular input. Streaming without detection and joint modeling of hands, objects, and bodies remain future work.

{
    \small
    \bibliographystyle{ieeenat_fullname}
    \bibliography{main}
}

\clearpage
\appendix
\setcounter{figure}{0}
\setcounter{table}{0}
\setcounter{equation}{0}
\renewcommand{\theHfigure}{supp.\arabic{figure}}
\renewcommand{\theHtable}{supp.\arabic{table}}
\renewcommand{\theHequation}{supp.\arabic{equation}}
\setcounter{secnumdepth}{0}
\setcounter{topnumber}{3}
\setcounter{totalnumber}{4}
\renewcommand{\topfraction}{0.92}
\renewcommand{\floatpagefraction}{0.72}
\renewcommand{\textfraction}{0.07}
\setcounter{dbltopnumber}{2}
\renewcommand{\dbltopfraction}{0.92}
\renewcommand{\dblfloatpagefraction}{0.6}

\section*{Appendix}

\subsection*{A. Teacher Pipeline Details}

The teacher first optimizes an initial 10 hours of in-the-wild video
for approximately 2000 steps; filtered outputs train the first ESTHER.
For later rounds, ESTHER forward inference supplies MANO initial estimates
on additional samples. The teacher refines them with the same objectives
for 200 steps, followed by the same geometric, reprojection, and human
filtering described below. Accepted teacher outputs expand the training
set. Deployment and testing use forward inference of the final ESTHER.

The 2D evidence is produced by an ensemble of detectors
(YOLO11-pose and ViTPose) run independently on both rectified pinhole views. Each detector
$d\in\{1,2\}$ returns, for every one of the 21 hand joints $j$, a 2D
location $\mathbf{u}_j^{d}$ and confidence $c_j^{d}$, and each
hand box a detector score $s^{d}$. Their outputs are fused using confidence, following WiLoR. With the valid set
$V_j=\{d:c_j^{d}\ge\tau\}$ (threshold $\tau{=}0.3$), the fused keypoint is
\begin{equation}
\hat{\mathbf{u}}_j=
\begin{cases}
\dfrac{c_j^{1}\mathbf{u}_j^{1}+c_j^{2}\mathbf{u}_j^{2}}{c_j^{1}+c_j^{2}},
& |V_j|=2 \ \text{and}\ \lVert\mathbf{u}_j^{1}-\mathbf{u}_j^{2}\rVert\le\epsilon\,w_b,\\[8pt]
\mathbf{u}_j^{d^{\star}}, & \text{otherwise,}
\end{cases}
\end{equation}
where $d^{\star}=\arg\max_{d\in V_j}c_j^{d}$ (near ties broken by the box
score $s^{d}$), $w_b$ is the hand box side, and joints with
$V_j=\emptyset$ are dropped. The agreement radius is $\epsilon{=}0.1$ (two
detections are averaged only if they fall within $10\%$ of the hand box
side of each other); it is set to about one MANO joint spacing at the
working hand scale and the fusion is insensitive to it in the
$0.08$--$0.15$ range. In words: two detectors that agree
are averaged with confidence weights, a lone confident detector is trusted,
and disagreements fall back to the estimate with higher confidence, retaining
each detector's reliable keypoints while suppressing the isolated false
positives that either produces alone. The stereo filtering stage then
removes geometrically implausible detections in two steps. First, joints whose triangulated depth falls outside the
$0.1$--$1.0$\,m distance range between hand and camera typical of head-mounted capture are
rejected. Second, a structural constraint on the relative distances between
wrist and finger joints removes isolated matches whose reconstructed bone
lengths deviate by more than $40\%$ from the canonical MANO template. Surviving keypoints and coarse 3D anchors form the
observation set $\mathcal{V}$ of the fitting objective
\begin{equation}
\begin{aligned}
\mathcal{L}_{\mathrm{fit}} =
&\lambda_{2d}\mathcal{L}_{2d}
+\lambda_p \mathcal{R}_{\mathrm{pose}}(\boldsymbol{\theta})\\
&\quad
+\lambda_b \|\boldsymbol{\beta}-\boldsymbol{\beta}_0\|_2^2
+\lambda_t \|\mathbf{t}-\mathbf{t}_0\|_2^2 ,
\end{aligned}
\end{equation}
where $\mathcal{L}_{2d}$ is a robust reprojection loss between projected
MANO joints and the filtered stereo keypoints over both views. Triangulation
serves as both the initialization and a soft constraint: the
triangulated wrist depth $\hat{z}_{\mathrm{wrist}}$ seeds the global
translation $\mathbf{t}_0$, and the mild
$\lambda_t\lVert\mathbf{t}-\mathbf{t}_0\rVert_2^2$ term then anchors the
solution loosely to it while $\mathcal{L}_{2d}$ drives the fit. The weight
$\lambda_t$ is small, so this is a soft prior the reprojection can override,
not a hard depth constraint. For the initial 10-hour seed set, WiLoR
initialization fuses MANO estimates from the left and right views in the calibrated
stereo frame. In subsequent labeling rounds, ESTHER predictions replace
this initialization; the optimization objectives and filtering remain the same.

The optimization runs in two stages. Stage~1 (shape lock) selects the 64
frames with the highest stereo detection confidence, optimizes
$(\boldsymbol\beta,\boldsymbol\theta,\mathbf{t})$ jointly on them under
$\mathcal{L}_{\mathrm{fit}}$, and averages the framewise shapes into one $\bar{\boldsymbol\beta}$ per hand. Stage~2 reoptimizes pose and
wrist per frame with $\bar{\boldsymbol\beta}$ fixed, then applies motion
optimization over 16-frame sliding windows with the objective
\begin{equation}
\begin{split}
\mathcal{L}_{\mathrm{motion}}
={}&\sum_{t}\Big[\lambda_{2d}\,\mathcal{L}_{2d,t}
+\lambda_{z}\,\rho\!\big(t_{z}^{(t)}-\hat{z}_{\mathrm{wrist}}^{(t)}\big)\\
&\quad+\lambda_{\omega}\,\rho\!\big(\angle(\boldsymbol\phi_{t+1},\boldsymbol\phi_{t})-\omega_{\max}\big)\\
&\quad+\lambda_{v}\,\rho\!\big(\lVert\mathbf{t}_{t+1}-\mathbf{t}_{t}\rVert-v_{\max}\big)\Big]\\
&+\lambda_{m}\,\mathcal{R}_{\mathrm{motion}}
\big(\boldsymbol\theta_{t:t+15}\big),
\end{split}
\end{equation}
where $\boldsymbol\phi_t$ is the global (wrist) orientation, i.e.\ the root
axis-angle component of the pose $\boldsymbol\theta_t$ itself (its last three
entries in the MANO parameterization); the angular penalty thus acts on a
subvector of $\boldsymbol\theta_t$, and $\angle(\cdot,\cdot)$ is the geodesic
angle between consecutive orientations. As in Stage~1, the triangulated
wrist depth $\hat{z}_{\mathrm{wrist}}^{(t)}$ enters only as the same
soft anchor on depth $t_z^{(t)}$ at each frame (small weight
$\lambda_z$, robust $\rho$); the reprojection term can override it, so it is
a prior rather than a hard depth constraint.
Here, as in $\mathcal{L}_{2d}$, $\rho$ is the Huber function (quadratic
within a small threshold $\delta$, linear beyond), which downweights
single-frame outliers. It acts on the raw residual for the depth anchor and
on the excess over the limit for the two velocity terms, so the velocity
penalty is one-sided and active only above $\omega_{\max}$ and $v_{\max}$
($\omega_{\max}\!\approx\!300^\circ$/s and
$v_{\max}\!\approx\!1.5$\,m/s at 30\,fps); the orientation term carries the
larger weight
($\lambda_{\omega}>\lambda_{v}$), because over a short 16-frame window the
global translation $\mathbf{t}$ is partly confounded by camera motion and is
therefore constrained more loosely. $\mathcal{R}_{\mathrm{motion}}$ is the
energy of a 16-frame motion prior, from an encoder pretrained on AMASS, over
the pose trajectory $\boldsymbol\theta_{t:t+15}$; it scores the orientation
and articulation dynamics of the hand and is independent of the global
translation. The smoothing weights are derived from the
prior's gradient feedback and from differences between frames, so windows with
implausible dynamics receive stronger regularization while genuinely fast
motion is preserved.

The pipeline uses stereo only where it is geometrically reliable and lets
the hand model and priors regularize unobserved structure: under object occlusion or self-occlusion the desired annotation is a structured
articulated state with semantic joints, MANO shape, and global wrist
translation, not the depth of whatever surface happens to be visible. This
suits egocentric capture, where hand size, camera distance, and
motion centered at the wrist occupy a constrained range.

The filtering above (confidence, depth range, plausibility of bone lengths) is the
coarse first stage: it prescreens the input 2D evidence before fitting.
Fitted sequence labels pass two further stages before entering
ESTHER3D. Stage two is a fine 2D check: it reprojects each fitted hand into both
views and computes the residual $r$, the mean pixel distance to the fused
keypoints over all joints of both views with confidence above $\tau$;
hand frames with $r>15$\,px are discarded. Stage three is
human review for the residual failures the automatic stages cannot catch (wrong
hand identity, implausible articulation). This coarse
(bone length/depth)~$\rightarrow$~fine (2D reprojection)~$\rightarrow$~human
screen is the filter described in the main paper.

\subsection*{B. Loss Definitions and Weights}

Training proceeds in the two stages of the main paper: the stereo image
encoder is first pretrained under its geometric guidance, then the
full model is trained jointly.

\textbf{Stage 1: image encoder pretraining.} The image encoder
is trained alone under the geometric guidance group
$\mathcal{L}_{\mathrm{img}}
=\lambda_{2d}\mathcal{L}_{2d}
+\lambda_{hm}\mathcal{L}_{hm}
+\lambda_{tri}\mathcal{L}_{tri}$
with $\lambda_{2d}{=}0.5$, $\lambda_{hm}{=}1.0$, $\lambda_{tri}{=}1.0$.
The 2D loss is a Huber loss on coordinates normalized by image size, masked by
joint visibility in each view; a blanked view contributes no 2D or heatmap
supervision. The heatmap loss is a cross entropy against Gaussian soft
targets on the prediction grid. The triangulation loss is an L1 loss between
$\mathbf{X}^{tri}$ and metric 3D joints, applied only to joints visible in
both views. Visibility uses binary cross entropy on valid image projections; depth uses cross entropy over logarithmically spaced bins.

\textbf{Stage 2: joint training of the full model.} The image encoder
(keeping its guidance from Stage 1) is joined by the motion encoder and fusion
decoder and trained end-to-end. The decoder output is mapped through the MANO
forward kinematics layer, and $\mathcal{L}_{\mathrm{FK}}$ combines L1 losses
on joints in the reference frame and canonical (wrist-relative) joints, an L1 mesh
vertex loss, a geodesic rotation loss and 6D consistency term on pose, an L1
shape loss, an L1 wrist translation loss, and articulated priors (PCA pose
prior, prior on joint limits, and shape regularization), with weights
$\lambda_{J}{=}1.0$, $\lambda_{Jcan}{=}1.0$, $\lambda_{V}{=}0.5$,
$\lambda_{\theta}{=}1.0$, $\lambda_{6d}{=}0.2$, $\lambda_{\beta}{=}0.01$,
$\lambda_{t}{=}2.0$, and prior weights $0.1/0.05/0.02$. In the same stage the
motion encoder is supervised at masked positions, reconstructing canonical 3D
joints (L1, per masked joint), MANO pose (L1, per masked frame), and root
velocity (L1 on frame differences), summed as $\mathcal{L}_{\mathrm{mot}}$
with weight $\lambda_{mot}{=}0.5$; the on-screen term
$\mathcal{L}_{\mathrm{on}}$ is a binary cross entropy against framewise labels indicating hand visibility in at least one retained view,
with weight $\lambda_{on}{=}0.5$.

\subsection*{C. Masking and Augmentation Recipe}

Token masking rates: individual joints $0.20$, entire frames $0.10$, contiguous temporal
span $0.10$ (span length 3), entire views $0.05$ (the first three via the
motion mask token, the last via the view mask token); anchor frames (every
$T/4$) are kept fully observed. Masked 2D observations always mask their
associated ray information. Image augmentation: Gaussian noise in each view with
standard deviation up to $0.08$; random erasing with probability $0.5$ (two
rectangles covering 10--40\% of each side); blanking entire views with
probability $0.12$; blanking entire frames with probability $0.08$, which also
provides on-screen negatives.

\subsection*{D. Camera Motion Flow Token}

The flow token is the dense classical optical flow of the whole crop between
two consecutive frames, encoded directly as a conditioning input. It uses a fast classical estimator over all crop pixels, including the hand. The estimator runs online for every frame and requires neither training nor ground truth.

The crops follow the hand: each frame's crop has its own intrinsic
$\mathbf{K}_t$ (its centre and scale track the detected hand every frame), so
even a static camera produces apparent motion purely from the moving crop
window. Using the crop intrinsics available at inference, we compensate
this before measuring flow: the previous crop is warped by the homography
$\mathbf{H}_t=\mathbf{K}_{t+1}\mathbf{K}_t^{-1}$ into the current crop's
coordinates, so displacement from crop motion cancels and the remaining flow
reflects real scene motion under the camera. Dense flow on this compensated
pair is downsampled by median pooling over an $8\times8$ grid, normalized by
the $224$-pixel crop side, with a flag indicating image validity for each cell:
\begin{equation}
\mathbf{f}_t=
\left[
\operatorname{vec}(\mathbf{F}_t/224),
\operatorname{vec}(\mathbf{M}_t)
\right]\in\mathbf{R}^{192},
\end{equation}
where $\mathbf{F}_t\in\mathbf{R}^{8\times8\times2}$ is the flow grid
($128$ values) and $\mathbf{M}_t\in\{0,1\}^{8\times8}$ ($64$ values) marks
cells within the frame, for $128{+}64{=}192$ dimensions; median pooling rejects flow
outliers.

The token carries both camera egomotion and hand motion without manually separating them. The token is fed to the motion encoder directly and, being a
conditioning input, is excluded from random token masking. The image and motion tokens, by contrast,
are masked during training; unable to lean on the masked appearance,
the encoder is forced to read the global, scene wide component of the flow as
camera motion and attribute the residual, hand localized component to
articulation. The decoupling of hand and camera motion is thus learned from
this always present flow cue, rather than absorbing head or crop motion into
finger pose.

\subsection*{E. Implementation Details}

The model dimension is 512 with 8 attention heads: 4 stereo transformer blocks, 4 temporal transformer blocks, and 2 fusion decoder layers over
16-frame clips. The full
model totals 354.5M parameters, of which the frozen DINOv3-L backbone
accounts for 303.1M; the trainable remainder is 51.3M (image encoder blocks
and heads 21.9M, motion encoder 12.8M, fusion decoder 16.6M). The decoder queries $\mathbf Q_p$, $\mathbf Q_s$, and $\mathbf Q_w$
represent pose, shape, and wrist, respectively. Each query gate is predicted
by an MLP from the query state, the pooled
mean of each source, and four evidence statistics: the mean and minimum
predicted visibility, the magnitude of the triangulated wrist, and the
triangulation spread. The DINOv3-L
backbone is frozen. The image encoder is pretrained, then the full
model is trained jointly with AdamW (learning rate $2\times10^{-4}$, cosine
schedule, 500 warmup steps, gradient clipping 1.0, bfloat16 autocast, losses
in fp32). Left hands are handled by mirroring the stereo rig, and predictions
are unmirrored for evaluation and projection. At deployment the detector is
used only to localize, crop, and match hands across stereo views;
the final ESTHER predicts MANO directly, without teacher optimization.

\subsection*{F. Camera Setups}

Our head-mounted RGB stereo rig is two synchronized color cameras on a rigid
headband with a $6.2$\,cm baseline, calibrated once per device. It is a
pinhole pair with focal length $624$\,px and principal point $(924,606)$ at $1920\times1200$ per eye. The network consumes these images directly.
The HOT3D devices instead carry strongly distorted wide-FOV
fisheye cameras: both Project Aria and Quest~3 use the Fisheye624 model
(FisheyeRadTanThinPrism: six radial, two tangential, and four
thin-prism coefficients), on monochrome streams
($640\times480$ for Aria, $1280\times1024$ for Quest~3). Table~\ref{supp:tab:cameras}
lists camera resolution, focal length, and baseline for each setup; full
intrinsics, distortion, and extrinsics are released with the benchmark.
Rectifying these fisheye views to pinhole is a design choice, so we ablate
several undistortion strategies (Appendix~M, Table~\ref{supp:tab:undist}); the
HOT3D numbers in the main paper use the strategy with the best average
accuracy, and ESTHER is in any case nearly invariant to the choice.

\textbf{Why rectify HOT3D to pinhole.} The HOT3D fisheye streams are
undistorted to a pinhole crop before they enter the model, for two reasons.
First, the 2D hand detectors (YOLO11-pose, ViTPose) are trained on
perspective imagery and localize markedly worse on curved fisheye frames;
rectification restores geometry of straight lines and thus reliable detection and
cropping. Second, a pinhole crop makes the Plucker ray tokens trivial to
compute, since each pixel projects back to a single straight ray through $\mathbf{K}^{-1}$, and matches our RGB rig's parameterization, so the same network consumes the HOT3D fisheye rigs unchanged.

\textbf{Undistortion and its costs.} For the Fisheye624 HOT3D cameras we
unproject each pixel with the released model and reproject onto a pinhole
camera, baking the
fisheye${\rightarrow}$undistort${\rightarrow}$crop${\rightarrow}$resize chain
into one crop intrinsic matrix $\mathbf{K}$. Rectifying a wide-FOV fisheye
to a pinhole plane is not free: the periphery is magnified, so edge pixels are
upsampled and lose effective resolution; the full field of view cannot map to
a finite pinhole image, so only a central FOV is kept and hands near the rim
can be cropped out; and the distortion model leaves a growing residual
reprojection error toward large incidence angles. Together with the grayscale
modality and different focal length, this is the domain gap that the small
HOT3D fine-tuning (main paper) closes.

\begin{table}[t]
\centering
\caption{Camera setups (camera resolution and focal length; stereo baseline). Fisheye rigs are rectified to pinhole
before use.}
\label{supp:tab:cameras}
\resizebox{\columnwidth}{!}{%
\begin{tabular}{@{}lcccc@{}}
\toprule
Setup & Resolution & Focal (px) & Baseline (mm) & Modality \\
\midrule
Collection rig (ours) & $1920\times1200$ & 623 & 62 & RGB \\
Test rig (mocap) & $1920\times1200$ & 623 & 62 & RGB \\
HOT3D Project Aria & $640\times480$ & 241 & 138 & monochrome \\
HOT3D Quest~3 & $1280\times1024$ & 504 & 64 & monochrome \\
\bottomrule
\end{tabular}
}
\end{table}

\subsection*{G. ESTHER3D Training Set: In-the-Wild Source Collection}

\begin{figure*}[t]
\centering
\includegraphics[width=0.80\textwidth]{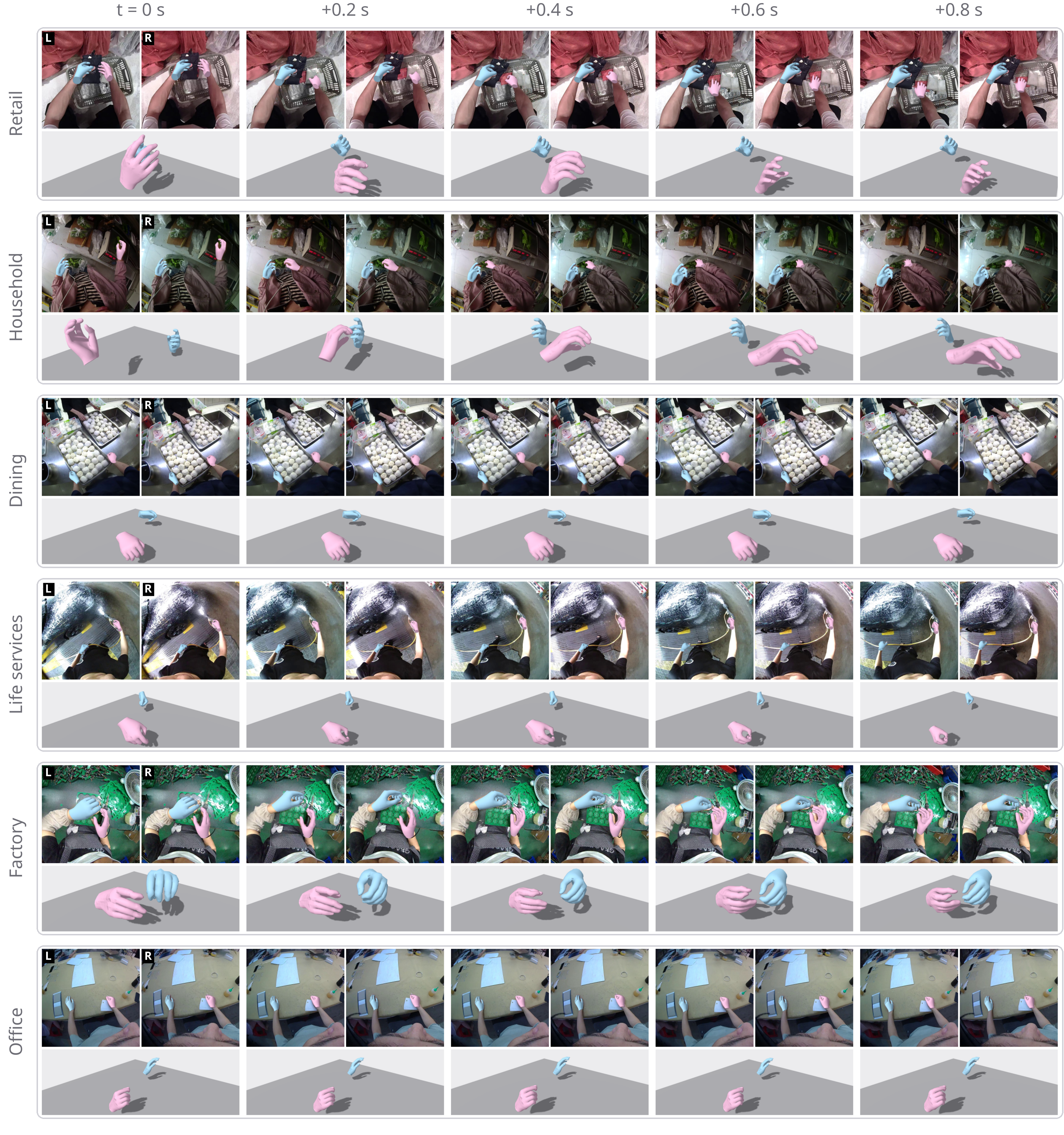}
\caption{ESTHER on in-the-wild ESTHER3D scenes across six industries.
Each block shows five timesteps of one scene (columns, spaced $0.2$\,s
apart); per timestep ESTHER's MANO meshes are projected into both stereo eyes
side by side (L for left and R for right, right hands in pink, left in blue)
with the same metric meshes shown in 3D below.}
\label{supp:fig:scenes}
\end{figure*}

The training set of ESTHER3D is the in-the-wild source collection, which comprises about 194 hours of
egocentric stereo video spanning six industry categories, 44 scene
categories, and 308 distinct real venues, with about 800 distinct
manipulation tasks performed by 121 collectors (main paper).
Below we list the scene categories and representative manipulation tasks per
industry, ordered by prevalence; Figure~\ref{supp:fig:scenes} shows examples of
these in-the-wild scenes.

\textbf{Retail} (44.0\,h; 11 scene categories, 44 venues).
Convenience stores dominate, followed by department stores, supermarkets,
clothing stores, stores selling daily goods, food and beverage retail, leather
fashion, hardware and tool shops, and toy stores. Representative tasks:
stocking and displaying merchandise, arranging bagged and boxed snacks,
hanging cable and accessory products, arranging bracelets and ornaments,
tidying shelves, folding and arranging garments, and cleaning the store.

\textbf{Household} (40.0\,h; 8 scene categories, 71 venues).
Rural homes with multiple stories, penthouse apartments, and apartments of diverse
layouts from one-bedroom to duplex. Representative tasks: cleaning floors,
making beds, organizing clothes and wardrobes, tidying kitchens, bedrooms,
and living rooms, arranging desktop items and sofa cushions, washing dishes,
wiping surfaces, and cooking.

\textbf{Dining} (32.0\,h; 8 scene categories, 57 venues).
Fast food restaurants in urban, village, and roadside settings, tea houses,
cafes and drink shops, dine-in restaurants and barbecue venues, and food
stalls. Representative tasks: arranging tables and chairs, cleaning counters
and table tops, sweeping and mopping, laying out tableware, clearing dishes,
washing kitchen containers, and sorting vegetables.

\textbf{Life services} (30.0\,h; 6 scene categories, 44 venues).
Auto repair shops, shops providing daily services, beauty salons, and electronics,
phone, and motorcycle repair shops. Representative tasks: removing and
installing car headlight parts, chassis repair, phone repair, applying
screen protectors, cleaning devices and counters, organizing tools, washing
hair, and washing towels.

\textbf{Factory} (26.0\,h; 5 scene categories, 42 venues).
Printing and book printing workshops, metalwork shops for tools and doors,
and wooden furniture workshops. Representative tasks: operating the main
workstation, cutting panels, grinding and polishing, repairing machinery and
press machines, positioning and punching, carrying materials, and packaging
profiles.

\textbf{Office} (22.0\,h; 6 scene categories, 50 venues).
Corporate office towers and government service centers. Representative
tasks: tidying office desks, arranging bookshelf items, tidying reception
coffee tables and tea sets, cleaning office areas, and clearing floor
litter.

\subsection*{H. ESTHER3D Test Set: Motion Capture Studio}

The test set of ESTHER3D is captured in a motion capture studio by the same
head-mounted stereo rig (calibrated with Kalibr, 6.2\,cm baseline) synchronized
with a Vicon system, giving true metric ground truth (MANO recovered from gloves/BVH hand skeletons plus raw marker trajectories). It totals about
10 hours from 6 subjects over 50 scripted manipulation tasks, distributed
across six instrumented scene zones that mirror the six training
industries: retail shelving (12 tasks), a life services corridor for walking/inspection (10), housekeeping (8), factory workbench (7), office reception (7),
and dining (6). Each zone uses textured backgrounds so the onboard visual
odometry stays stable, and every method is evaluated zero-shot.

\begin{figure*}[p]
\centering
\includegraphics[height=0.90\textheight]{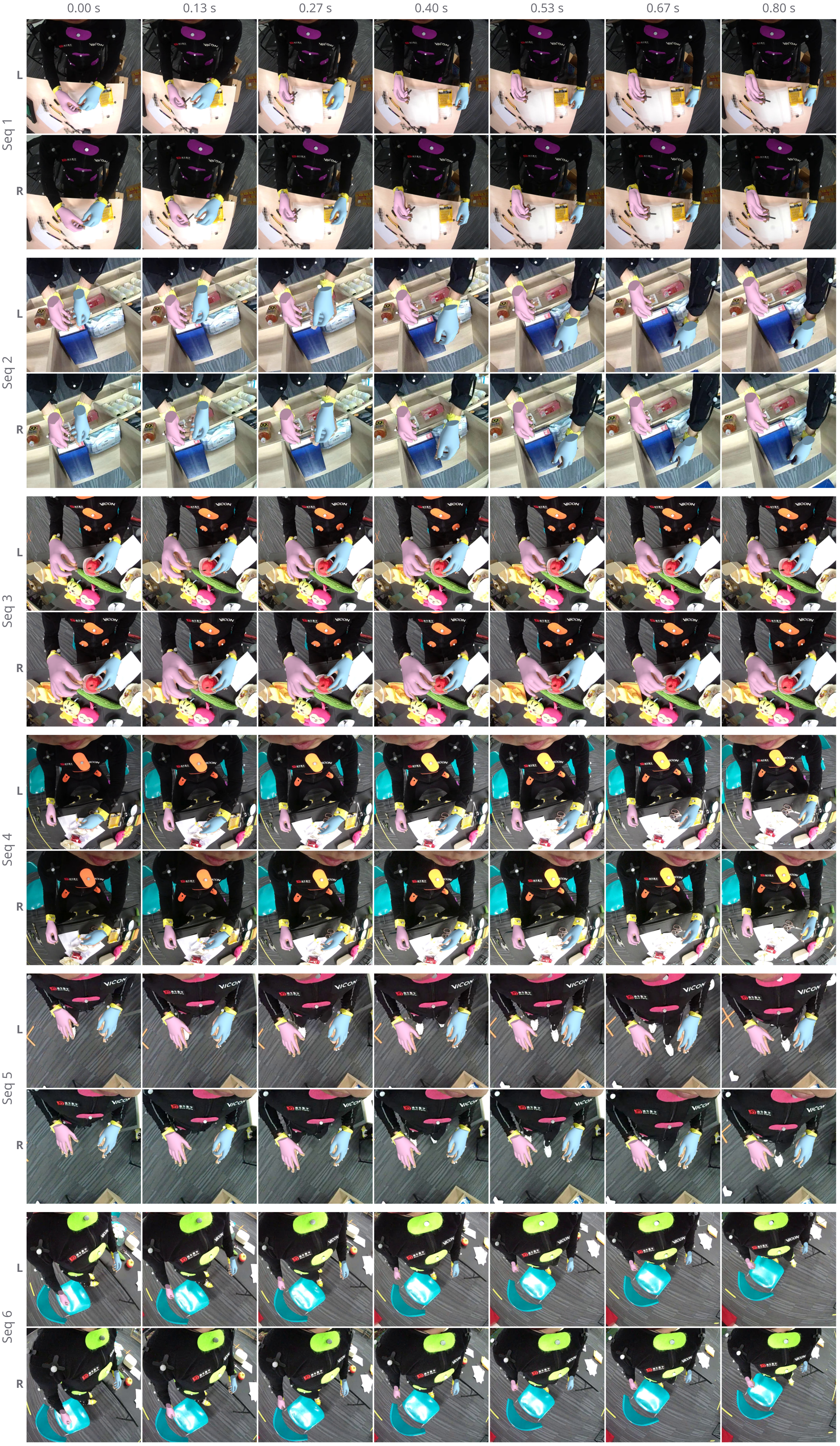}
\caption{Motion capture ground truth of the ESTHER3D test set. Six sequences
(Seq~1--6), shown in two rows for the left eye (L, cam0) and right eye (R, cam1), across seven timesteps (columns,
labeled by time). The ground truth hands (MANO fitted to the motion capture
hand joints) are projected into each view; they are the true metric
supervision against which all methods are evaluated zero-shot.}
\label{supp:fig:gtvis}
\end{figure*}

To cover manipulation broadly, the protocol is organized around a taxonomy of
14 elementary hand actions, and the 50 tasks are designed so that every action
is exercised. Each action maps to a concrete manipulation: wipe (surface and
equipment cleaning), grasp (picking goods
and tools), place (aligning and setting down), twist (turning screws and caps), fold/spread (folding garments, making beds), gather (clearing
and tidying), shelve (restocking, book returns), press (buttons and
switches), carry (two-handed lifting), flip (turning objects and pages),
point (guiding and reception gestures), bimanual (asymmetric bimanual assembly and packaging), pinch (picking small parts and screws), and
handover (passing an object to another person). Table~1 of the main
paper reports the accuracy of each method on this set.
Figure~\ref{supp:fig:gtvis} visualizes the motion capture ground truth used for
evaluation: MANO hand skeletons recovered from gloves and projected into the head-mounted camera across six test sequences and seven timesteps each.

\subsection*{I. Teacher Pipeline Validation}

We validate the teacher pipeline on the 10-hour motion capture test set of
Appendix~H, where it runs without access to the ground truth.

\textbf{Mapping ground truth into camera coordinates.} Predictions are compared to ground
truth directly in the camera frame, with no individual clip alignment. The
camera's pose in the mocap world has six degrees of freedom and comes from
a rigid body with four markers on the rig combined with the calibrated
extrinsic transform from the device to the camera; ground truth hands are mapped into the camera
via a fixed similarity solved from wrist markers (residual 10.4\,mm,
approximately the offset between joint and marker). Each link of the chain is
verified independently: forward kinematics bone lengths, joint
correspondence via a shuffle test, and time synchronization solved twice by
independent procedures: direct marker/detection alignment and a pose sweep, agreeing to 10\,ms. Evaluation covers both hands throughout the 10 hours of mocap test video.

The teacher's accuracy on this set (and the comparison of methods) is
reported in the main paper; here we only give the measurement chain above
and additional ground truth examples in Figure~\ref{supp:fig:gtvis}.

\subsection*{J. Statistical Rigor: Bootstrap Confidence Intervals}

The architectures in Table~1 of the main paper share frozen image
evidence, training data, forward kinematics losses, and budget. To assess
sampling variability, we bootstrap clip scores from the full 10-hour ESTHER3D motion capture
test set using $2000$ resamples
(Table~\ref{supp:tab:bootstrap}). ESTHER's updated MPJPE point estimate is
17.7\,mm; its corresponding bootstrap uncertainty is not reported here.
For jitter, ESTHER's $95\%$ interval $[3.10,3.52]$ lies below the best
baseline's $[7.98,9.17]$. These intervals quantify sampling variability for a fixed checkpoint, not variation across training seeds. The wrist
error ($21.3$\,mm) is on par with the best multiview methods. The
perturbation sweeps in Tables~2 and~3 of the main paper use this same model.

\begin{table}[t]
\centering
\caption{Bootstrap statistics on the full 10-hour ESTHER3D test set
($2000$ resamples): mean\,$\pm$\,std where available. Errors in mm,
jitter in m/s$^2$. ESTHER's MPJPE is the updated point estimate only;
its jitter $95\%$ interval does not overlap any baseline's.}
\label{supp:tab:bootstrap}
\resizebox{\columnwidth}{!}{%
\begin{tabular}{@{}lcccc@{}}
\toprule
Method & MPJPE $\downarrow$ & PA-MPJPE $\downarrow$ & Wrist $\downarrow$ & Jitter $\downarrow$ \\
\midrule
Hiera-FTL & $43.4\pm1.2$ & $10.0\pm0.2$ & $277\pm5$ & $79.5\pm3.6$ \\
Epipolar Trans. & $19.0\pm0.5$ & $9.0\pm0.1$ & $\mathbf{20.9\pm0.5}$ & $13.3\pm0.4$ \\
MVGFormer & $19.2\pm0.6$ & $8.9\pm0.1$ & $21.0\pm0.4$ & $8.7\pm0.3$ \\
POEM-v2 & $19.4\pm0.6$ & $8.9\pm0.1$ & $22.5\pm0.4$ & $8.5\pm0.3$ \\
\cellcolor{gray!12}ESTHER (ours) & \cellcolor{gray!12}$\mathbf{17.7}$ & \cellcolor{gray!12}$\mathbf{8.6\pm0.1}$ & \cellcolor{gray!12}$21.3\pm0.5$ & \cellcolor{gray!12}$\mathbf{3.3\pm0.1}$ \\
\bottomrule
\end{tabular}}
\end{table}

\subsection*{K. Robustness to Failure Modes}

\begin{figure*}[t]
\centering
\includegraphics[width=0.98\textwidth]{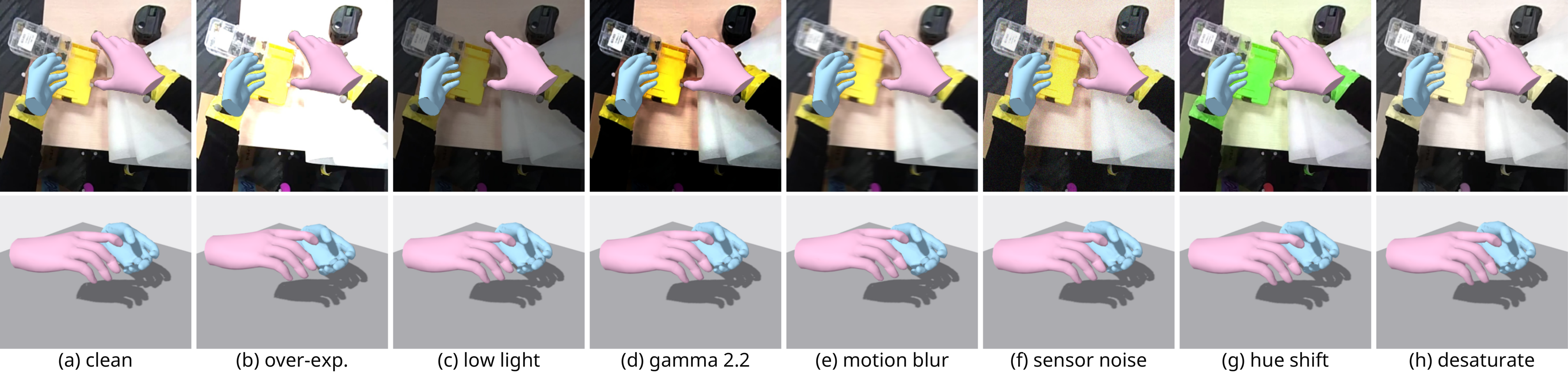}
\caption{ESTHER under input corruptions on an ESTHER3D motion capture clip:
MANO meshes projected on the degraded left view (top) and shown in 3D (bottom),
for the clean input (a) and seven corruptions (b--h) spanning lighting, blur,
noise, and color shifts. The mesh stays stable throughout.}
\label{supp:fig:robust}
\end{figure*}

Beyond the missing views and dropped frames of the main paper, we stress
ESTHER with seven photometric and temporal corruptions applied to the
ESTHER3D test frames, spanning the main egocentric failure modes: overexposure, underexposure, and nonlinear gamma (lighting), motion
blur, sensor noise, a hue shift (skin tone / color), and desaturation
(Table~\ref{supp:tab:robust}, Figure~\ref{supp:fig:robust}). The gloved appearance
gap is the test condition itself (Appendix~H): ESTHER is trained on bare
in-the-wild hands, yet every number here is measured on the gloved
motion capture subject, so the shift from bare hands to gloves is already priced
in.

Across all seven corruptions the metrics barely move: MPJPE stays within
$0.5$\,mm of the clean $17.7$ (worst case $18.2$ under sensor noise),
PA-MPJPE within $0.2$\,mm, wrist within about $2$\,mm of the clean $21.3$,
and jitter within approximately $0.4$\,m/s$^2$. Frozen DINOv3
evidence and heavy image augmentation during training make the encoder
largely invariant to appearance, and the temporal motion encoder absorbs blur
as it absorbs dropped frames. Skin tone (hue) and illumination shifts in
particular leave MPJPE unchanged at 17.7\,mm, matching clean input.

\begin{table}[t]
\centering
\caption{ESTHER under input corruption on the ESTHER3D test set. Errors in
mm, jitter in m/s$^2$; MPJPE under corrupted input is within 0.5\,mm
of the clean condition.}
\label{supp:tab:robust}
\resizebox{\columnwidth}{!}{%
\begin{tabular}{@{}lcccc@{}}
\toprule
Corruption & MPJPE & PA-MPJPE & Wrist & Jitter \\
\midrule
\cellcolor{gray!12}clean & \cellcolor{gray!12}$17.7$ & \cellcolor{gray!12}$8.60$ & \cellcolor{gray!12}$21.3$ & \cellcolor{gray!12}$3.31$ \\
overexposure ($\times1.6$) & $17.6$ & $8.66$ & $20.1$ & $3.71$ \\
low light ($\times0.5$) & $17.7$ & $8.57$ & $21.5$ & $3.34$ \\
gamma $2.2$ & $17.7$ & $8.49$ & $21.9$ & $3.33$ \\
motion blur ($k{=}15$) & $18.0$ & $8.66$ & $19.5$ & $3.24$ \\
sensor noise ($\sigma{=}25$) & $18.2$ & $8.68$ & $21.6$ & $3.73$ \\
hue shift ($+25$, skin tone) & $17.7$ & $8.60$ & $21.9$ & $3.37$ \\
desaturate ($\times0.4$) & $17.8$ & $8.64$ & $22.1$ & $3.39$ \\
\bottomrule
\end{tabular}}
\end{table}

\subsection*{L. On-Screen Gating and Jitter Protocol}

\textbf{Pose metrics.} MPJPE measures wrist-relative joint error;
PA-MPJPE additionally applies Procrustes alignment. MPJPE-absolute denotes
joint error without wrist alignment. The ablations in the main paper use MPJPE
on the ESTHER3D test set, with full ESTHER at 17.7\,mm.

\textbf{Jitter definition.} Jitter is the mean second temporal difference of
the joint positions, $\lVert\mathbf{S}_{t-1}-2\mathbf{S}_{t}+\mathbf{S}_{t+1}\rVert$
averaged over the $21$ joints and scaled by $\mathrm{fps}^2$
($\mathrm{fps}=30$), taken only over triplets whose three frames are all
on-screen. No acceleration is formed across an off-screen frame or an
on/off boundary, so a hand that is gated off contributes nothing to jitter.

\textbf{On-screen gating and off-screen frames.} A hand is on-screen
(ground truth) when at least half of its $21$ joints project inside the
image in at least one eye; ESTHER predicts hand visibility probabilities
that we threshold at $0.5$. Pose, wrist, and jitter are scored only on
frames where the hand is visible. Frames without a visible hand have no valid target, so gating determines which frames enter each metric.

\textbf{Gate accuracy and its effect on jitter.} On the full 10-hour motion capture test
set the gate is exact: precision, recall, and $\mathrm{F}_1$ are all $1.00$
over all evaluated hand frames (no false positives or negatives), so jitter over all frames and with ground truth or predicted gating coincides at $3.31$\,m/s$^2$; the reported value therefore does not depend on the gating protocol here. To probe
gate errors we run a stress test with the hand outside the view that blanks the middle four
frames of every clip. The gate flags them off-screen (recall $1.00$);
jitter gated by predictions holds at $3.35$\,m/s$^2$, whereas ungated
jitter rises to $4.61$ because the on/off/on transition injects a spurious
acceleration spike. A missed off-screen frame (a false negative) would thus
inflate jitter through such boundary spikes; the gate is what keeps the
metric clean.

\subsection*{M. HOT3D Robustness to Undistortion Strategy}

Because the HOT3D cameras are wide-FOV Fisheye624 (Appendix~F), the
rectified pinhole crop fed to the network is a design choice, and a natural
question is how much the results depend on it. On a held-out HOT3D split we
rerun every method under four undistortion strategies: axis-aligned
wide-pinhole undistortion at a virtual focal of 120\,px (offbb120, our
default), the same pipeline at 90\,px (narrow) and 150\,px (wide), and a
ray-aligned adaptive-focal hand crop with minimal black border (hc).
This is not a direct accuracy comparison: ESTHER is trained
on the offbb120 rectification and the multiview baselines use their released weights, which would make an absolute ranking unfair. What
Table~\ref{supp:tab:undist} measures instead is each method's stability:
how much its own numbers move when only the rectification changes, all else
fixed.

ESTHER is essentially invariant to the rectification choice. Its MPJPE stays within $32.3$--$32.5$\,mm and its jitter within
$16.3$--$16.8$\,m/s$^2$ across the three offbb focals, including the
90 and 150\,px settings it never saw in training. In contrast, the multiview baselines vary substantially: POEM's wrist error moves
$656\!\rightarrow\!450\!\rightarrow\!928$\,mm across the focal sweep alone,
and the methods based on anchors only recover on the hc strategy that crops the hand directly. This follows from direct absolute wrist prediction:
ESTHER's decoded wrist does not depend on a triangulated anchor whose
geometry shifts with the rectification, so the reconstruction is stable under
the undistortion pipeline rather than tuned to one.

\begin{table}[t]
\centering
\caption{HOT3D robustness to undistortion strategy on a held-out split.
Each cell: MPJPE / wrist (mm) / jitter (m/s$^2$). The comparison measures stability across strategies:
ESTHER barely moves (including on the 90/150\,px focals it was not trained
on) while the multiview baselines swing widely.}
\label{supp:tab:undist}
\resizebox{\columnwidth}{!}{%
\begin{tabular}{@{}lcccc@{}}
\toprule
Method & offbb120 & offbb90 & offbb150 & hc \\
\midrule
\cellcolor{gray!12}ESTHER (ours) & \cellcolor{gray!12}\textbf{32.4 / 20.8 / 16.3} & \cellcolor{gray!12}32.5 / 22.8 / 16.6 & \cellcolor{gray!12}32.3 / 23.6 / 16.8 & \cellcolor{gray!12}28.9 / 48.0 / 15.6 \\
POEM-v2 & 101.3 / 656 / 635 & 100.9 / 450 / 395 & 103.6 / 928 / 1044 & 46.3 / 65.7 / 48 \\
MVGFormer & 95.9 / 429 / 449 & 95.9 / 291 / 265 & 97.6 / 633 / 767 & 43.1 / 69.9 / 52 \\
Epipolar Trans. & 69.4 / 414 / 318 & 70.6 / 303 / 191 & 69.9 / 571 / 541 & 43.2 / 66.9 / 66 \\
Hiera-FTL & 80.9 / 401 / 105 & 78.7 / 390 / 100 & 82.8 / 435 / 118 & 78.1 / 377 / 91 \\
\bottomrule
\end{tabular}}
\end{table}

\subsection*{N. Label Quality Analysis}

The motion capture test set measures the labeling procedure against true
metric ground truth, but only in the studio. Here we report 2D agreement
and geometric sensitivity to first order of the in-the-wild labels.

\textbf{Residual.} For every labeled hand frame we project the 21 label joints
into both rectified pinhole views through the calibrated rig and compare them
with the fused 2D keypoints of Appendix~A (YOLO11-pose and ViTPose, detected on
the same views); the residual $r$ is the mean pixel distance over all joints of
both views whose confidence exceeds $\tau{=}0.3$. This is exactly the quantity
of the fine 2D check (Appendix~A), which rejects hand frames with $r>15$\,px. The
statistics cover the in-the-wild training set of ESTHER3D (Appendix~G).

\textbf{From 2D residual to 3D error.} A residual of $r$ pixels at hand depth
$z$ corresponds to a displacement parallel to the image plane of $rz/f$, the
component of the 3D error that 2D observes directly. The component along the
viewing ray is only constrained through the disparity between the two views:
a disparity error $\delta d$ shifts the triangulated depth by
$\delta z=z^2\delta d/(fb)$. With independent errors in the two views
$\delta d\approx\sqrt{2}\,r$; if both views err in opposite directions,
$\delta d\le 2r$. Table~\ref{supp:tab:labelq} reports distributions of these quantities over hand frames.

\begin{table}[h]
\centering
\caption{In-the-wild label quality from 2D (hand frames passing the 15\,px
check). Depth $z$ is the label wrist depth.}
\label{supp:tab:labelq}
\resizebox{\columnwidth}{!}{%
\begin{tabular}{@{}lccc@{}}
\toprule
Quantity & Median & Mean & p90 \\
\midrule
Reprojection residual $r$ (px) & 5.3 & 5.8 & 8.6 \\
Hand depth $z$ (m) & 0.32 & 0.33 & 0.46 \\
In-plane discrepancy $rz/f$ (mm) & 2.8 & 3.0 & 4.6 \\
Depth per pixel of disparity $z^2/(fb)$ (mm) & 2.8 & 3.2 & 5.6 \\
Depth sensitivity, independent views (mm) & 21.3 & 25.1 & 44.2 \\
Depth sensitivity, opposing view errors (mm) & 30.1 & 35.5 & 62.5 \\
\bottomrule
\end{tabular}}
\end{table}

94.6\% of the hand frames pass the 15\,px check and 94.6\% of those lie below
10\,px.

\textbf{Interpretation.} The approximately 3\,mm in-plane discrepancy and
21\,mm depth sensitivity summarize reprojection agreement and error
propagation under the stated assumptions. They do not bound true 3D label
error, since the detectors used for fitting can share errors with the
labels. The 21\,mm quantity assumes independent view errors; opposing view
errors give the separate 30.1\,mm median in Table~\ref{supp:tab:labelq}.
Metric accuracy is measured against motion capture: teacher MPJPE is
15.9\,mm and wrist error is 19.1\,mm, as reported in the main paper.

\subsection*{O. Ethics Statement}

Participants in both collections gave informed consent before recording
and were compensated. Consent covered model training, evaluation, and
release as an anonymized benchmark for noncommercial research. The
recordings contain hands and manipulated objects/workspaces, with no
personally identifying facial imagery.

\end{document}